\documentclass[12pt]{article}

\usepackage{amssymb,bm,amsthm}
\usepackage{enumerate}
\usepackage{color, colortbl}
\usepackage{booktabs, multirow}
\definecolor{darkgreen}{rgb}{0, 0.5, 0}
\definecolor{red}{rgb}{1, 0, 0}
\definecolor{purple}{rgb}{0.5, 0, 0.5}
\usepackage{chngpage}
\usepackage{comment}
\usepackage{mfirstuc}
\usepackage{mathtools}
\usepackage{lettrine}
\usepackage{pxfonts}
\usepackage{subfig}

\newcommand\eg{\textit{e.g.}}
\newcommand\ie{\textit{i.e.}}

\newcommand\iid{\textit{i.i.d.}}

\newcommand\doubleP{\mathbb{P}}

\newcommand{\norm}[1]{\left\lVert#1\right\rVert}

\newcommand{\beq}{\begin{equation}}
\newcommand{\eeq}{\end{equation}}
\newcommand{\beqnn}{\begin{equation*}}
\newcommand{\eeqnn}{\end{equation*}}
\newcommand{\beqy}{\begin{eqnarray}}
\newcommand{\eeqy}{\end{eqnarray}}
\newcommand{\beqynn}{\begin{eqnarray*}}
\newcommand{\eeqynn}{\end{eqnarray*}}
\newcommand{\bit}{\begin{itemize}}
\newcommand{\eit}{\end{itemize}}
\newcommand{\ben}{\begin{enumerate}}
\newcommand{\een}{\end{enumerate}}
\newcommand{\bex}{\begin{example}}
\newcommand{\eex}{\end{example}}

\newcommand{\balg}[1]{\begin{algorithm} \caption{#1}}
\newcommand{\ealg}{\end{algorithm}}

\newcommand{\balgc}{\begin{algorithmic}[1]}
\newcommand{\ealgc}{\end{algorithmic}}

\newcommand{\bary}{\begin{array}}
\newcommand{\eary}{\end{array}}
\newcommand{\bmx}{\begin{bmatrix}}
\newcommand{\emx}{\end{bmatrix}}
\newcommand{\bsmx}{\left[\begin{smallmatrix}}
\newcommand{\esmx}{\end{smallmatrix}\right]}
\newcommand{\bmxc}[1]{\left[\begin{array}{@{}#1@{}}}
\newcommand{\emxc}{\end{array}\right]}
\newcommand{\bcn}{\begin{center}}
\newcommand{\ecn}{\end{center}}

\newcommand{\diag}{\mathrm{diag}}

\newcommand{\trace}{\mathrm{trace}}

\newcommand{\Rbb}{{\mathbb{R}}}

\newcommand{\x}{{\boldsymbol{x}}}

\newcommand{\z}{\boldsymbol{z}}

\providecommand{\norm}[1]{\lVert#1\rVert}

\providecommand{\abs}[1]{| #1 |}

\newenvironment{theorem}[2][Theorem]{\begin{trivlist}
		\item[\hskip \labelsep {\bfseries #1}\hskip \labelsep {\bfseries #2.}]}{\end{trivlist}}

\usepackage{fullpage}
 
\usepackage{cite}
\usepackage{amsmath,amssymb,amsfonts}
\usepackage{times}
\usepackage{epsfig}
\usepackage{graphicx}
\usepackage{CJK}
\usepackage{algorithmic}
\usepackage{graphicx}
\usepackage{diagbox}
\usepackage{floatrow}
\usepackage[table]{xcolor}
\usepackage{wrapfig}
\usepackage[margin=1in,letterpaper]{geometry} 
\usepackage{textcomp}
\def\BibTeX{{\rm B\kern-.05em{\sc i\kern-.025em b}\kern-.08em
    T\kern-.1667em\lower.7ex\hbox{E}\kern-.125emX}}
\makeatletter
\let\NAT@parse\undefined
\makeatother
\usepackage{hyperref}
\newtheorem{definition}{Definition}
\newcommand*{\rom}[1]{\expandafter\@slowromancap\romannumeral #1@}
\newfloatcommand{capbtabbox}{table}[][\FBwidth]

\hypersetup{
   pdftitle={Sitao proposal},
   colorlinks=true,
   citecolor=blue,
   linkcolor=blue,
   urlcolor=blue
}
\begin{document}
	\title{On Addressing the Limitations of Graph Neural Networks}
     \author{
     Sitao Luan$^{1,2}$\\
     \{sitao.luan@mail\}.mcgill.ca\\
     $^1$McGill University; $^2$Mila\\
     }
	\maketitle

\begin{abstract}
This report gives a comprehensive summary of two problems about graph convolutional networks (GCNs): over-smoothing and heterophily challenges, and outlines future directions to explore. 
\end{abstract}

\section{Introduction}
Many real-world problems can be modeled as graphs. 
Recently, neural network based approaches have achieved significant progress for solving large, complex, graph-structured problems \cite{hamilton2017inductive, kipf2016classification, liao2019lanczos, gilmer2017neural, monti2017geometric, defferrard2016fast, hua2022graph}. Inspired by the success of Convolutional Neural Networks (CNNs) \cite{lecun1998gradient} in computer vision \cite{li2018adaptive}, graph convolution defined on the graph Fourier domain stands out as the key operator and one of the most powerful tools for using machine learning to solve graph problems. Although with high expressive power, GCNs still suffer from several difficulties, \eg{} the over-smoothing problem limits deep GCNs to sufficiently exploit multi-scale information, heterophily problem makes the graph-aware models underperform the graph-agnostic models. This report summarizes the methods we have proposed to address those challenges and puts forward some research problems we will investigate.

To fully explain the above problems, in subsection \ref{sec:notation_background}, we will first introduce the notation and background knowledge of graph networks. In section \ref{sec:deep_gnns}, we introduce the loss of expressive power of deep graph neural networks (GNNs)  and propose snowball and truncated Krylov architecture to address it; in section \ref{sec:heterophily}, we analyze heterophily problems for the existing GNNs and propose Adaptive Channel Mixing (ACM) architecture to address it.

\paragraph{Main Contribution} In section \ref{sec:deep_gnns}, we first point out that the output of deep GCN with ReLU activation function will suffer from loss of rank problem under certain conditions and this can cause deep GCN lose expressive power. We then prove that Tanh is better at preserving the rank of the output and verify this claim with numerical tests. Then we find a way to deepen GCN in block Krylov form and propose snowball and truncated Krylov networks which perform better than state-of-the-arts (SOTA) model on semi-supervised node classification tasks on 3 benchmark datasets. Besides, we point out that finding \textbf{a specifically tailored weight initialization scheme for GCNs} can be an promising direction to address over-smoothing efficiently in section \ref{sec:remaining_oversmoothing}. In section \ref{sec:heterophily}, we first illustrate the insufficiency of the current homophily metrics and propose aggregation homophily based on a new similarity matrix. We then show the advantage of the new homophily metric over the existing ones on synthetic graph. Based on the similarity matrix, we define diversification distinguishability of a node and demonstrate why high-pass filters can help to address heterophily problem. To include both low-pass and high-pass in GNNs, we extend filterbank method and propose ACM and ACMII frameworks that can boost the performance of baseline GNNs on heterophilous graphs.


\subsection{Notation and Background Knowledge}
\label{sec:notation_background}
Suppose we have an undirected graph $\mathcal{G}=(\mathcal{V},\mathcal{E}, A)$, where $\mathcal{V}$ is the node set with $\abs{\mathcal{V}}=N$; $\mathcal{E}$ is the edge set without self-loop; $A \in \mathbb{R}^{N\times N}$ is the symmetric adjacency matrix with $A_{ij}=1$ if and only if $e_{ij} \in \mathcal{E}$, otherwise $A_{ij}=0$; $D$ is the diagonal degree matrix, \ie{} $D_{ii} = \sum_j A_{ij}$ and $\mathcal{N}_i=\{j: e_{ij} \in \mathcal{E}\}$ is the neighborhood set of node $i$. A graph signal is a vector $\bm{x} \in \mathbb{R}^N$ defined on $\mathcal{V}$, where $x_i$ is defined on the node $i$. We also have a feature matrix ${X} \in \mathbb{R}^{N\times F}$ whose columns are graph signals and each node $i$ has a corresponding feature vector ${X_{i:}}$ with dimension $F$, which is the $i$-th row of ${X}$. We denote $Z\in \mathbb{R}^{N\times C}$ as label encoding matrix, where $Z_{i:}$ is the one hot encoding of the label of node $i$ and $C$ is the total number of classes.

The (combinatorial) graph Laplacian is defined as $L = D - A$, which is a Symmetric Positive Semi-Definite (SPSD) matrix \cite{chung1997spectral}. Its eigendecomposition gives $L=U\Lambda U^T$, where the columns of $U\in \Rbb^{N\times N}$ are orthonormal eigenvectors, namely the \textit{graph Fourier basis}, $\Lambda = \diag(\lambda_1, \ldots, \lambda_N)$ with $\lambda_1 \leq \cdots \leq \lambda_N$, and these eigenvalues are also called \textit{frequencies}. The graph Fourier transform of the graph signal $\x$ is defined as $\bm{x}_\mathcal{F} = U^{-1} \bm{x} = U^{T} \bm{x}=[\bm{u}_1^T\x, \ldots, \bm{u}_N^T\x]^T$, where $\bm{u}_i^T \bm{x}$ is the component of $\bm{x}$ in the direction of $\bm{u_i}$. 

Some graph Laplacian variants are commonly used, \eg{} the symmetric normalized Laplacian $L_{\text{sym}} = D^{-1/2} L D^{-1/2} = I-D^{-1/2} A D^{-1/2}$ and the random walk normalized Laplacian $L_{\text{rw}} = D^{-1} L = I - D^{-1} A$. The eigenvalues of $L_{\text{rw}}$ and $L_{\text{sym}}$ 
are the same and are in $[0,2)$, and their corresponding eigenvectors satisfy $\bm{u}_{\text{rw}}^i = D^{-1/2} \bm{u}_{\text{sym}}^i$. 

The affinity (transition) matrices can be derived from the Laplacians, \eg{} $A_\text{rw} = I - L_\text{rw} = D^{-1} A$, $A_\text{sym} = I-L_\text{sym} = D^{-1/2} A D^{-1/2}$.
Then  $\lambda_i(A_\text{rw}) = \lambda_i(A_\text{sym}) = 1- \lambda_i(A_\text{sym}) = 1- \lambda_i(A_\text{rw}) \in (-1,1]$. Renormalized affinity and Laplacian matrices are introduced in \cite{kipf2016classification}  as $\hat{A}_\text{sym} = \tilde{D}^{-1/2} \tilde{A}\tilde{D}^{-1/2}$, $\hat{L}_{\text{sym}} = I - \hat{A}_\text{sym}$, where $\tilde{A} \equiv A+I, \tilde{D} \equiv D+I$ and it essentially adds a self-loop and is widely used in Graph Convolutional Network (GCN) as follows:
\begin{equation}
    \label{eq:gcn_original}
   Y = \text{softmax} (\hat{A}_\text{sym} \; \text{ReLU} (\hat{A}_\text{sym} {X} W_0 ) \; W_1 )
\end{equation}
where $W_0 \in \Rbb^{F\times F_1}$ and $W_1 \in \Rbb^{F_1\times O}$ are parameter matrices. GCN can learn by minimizing the following cross entropy loss
\begin{equation}
\label{eq:cross_entropy_loss}
     \mathcal{L}  = -\trace(Z^T \log Y).
\end{equation}

The random walk renormalized matrix $\hat{A}_{\text{rw}} = \tilde{D}^{-1} \tilde{A}$ can also be applied to GCN and it has the same eigenvalues as $\hat{A}_{\text{sym}}$. The corresponding Laplacian is defined as $\hat{L}_{\text{rw}} = I - \hat{A}_\text{rw}$ Specifically, the nature of random walk matrix makes $\hat{A}_{\text{rw}}$ behaves as a mean aggregator $(\hat{A}_{\text{rw}} \bm{x})_i = \sum_{j\in\{\mathcal{N}_i \cup i\}} {x}_j/(D_{ii}+1)$ which is applied in \cite{hamilton2017inductive} and is important to bridge the gap between spatial- and spectral-based graph convolution methods.


\section{Loss of Expressive Power of Deep Graph Neural Networks}
\label{sec:deep_gnns}


One major problem of the existing GCNs is the low expressive power limited by their shallow learning mechanisms \cite{zhang2018graph, wu2019survey}. There are mainly two reasons why an architecture that is scalable in depth has not been achieved yet. First, this problem is difficult: considering graph convolution as a special form of Laplacian smoothing \cite{li2018deeper}, networks with multiple convolutional layers will suffer from an over-smoothing problem that makes the representation of even distant nodes indistinguishable \cite{zhang2018graph}. Second, some people think it is unnecessary: for example, \cite{bronstein2016geometric} states that it is not necessary for the label information to totally traverse the entire graph and one can operate on the multi-scale coarsened input graph and obtain the same flow of information as GCNs with more layers. Acknowledging the difficulty, we hold on to the objective of deepening GCNs since the desired compositionality\footnote{The expressive power of a sound deep Neural Network (NN) architecture should be expected to grow with the increment of network depth \cite{lecun2015deep, hinton2006fast}.} will yield easy articulation and consistent performance for problems with different scales.

In subsection \ref{sec:oversmoothing_tknet}, we first analyze the limits of deep GCNs brought by over-smoothing and the activation functions. Then, we show that any graph convolution with a well-defined analytic spectral filter can be written as a product of a block Krylov matrix and a learnable parameter matrix in a special form. Based on this, we propose two GCN architectures that leverage multi-scale information in different ways and are scalable in depth, with stronger expressive powers and abilities to extract richer representations of graph-structured data. For empirical validation, we test different instances of the proposed architectures on multiple node classification tasks. The results show that even the simplest instance of the architectures achieves state-of-the-art performance, and the complex ones achieve surprisingly higher performance.  In subsection \ref{sec:remaining_oversmoothing}, we propose to study an over-smoothing problem and give some ideas. 

\subsection{A Stronger Multi-scale Deep GNN with Truncated Krylov Architecture}
\label{sec:oversmoothing_tknet}
Suppose we deepen GCN in the same way as \cite{kipf2016classification, li2018deeper}, we have
\begin{equation}\label{eq1}
{Y} = \text{softmax} (\hat{A}_\text{sym}  \; \text{ReLU} ( \cdots \hat{A}_\text{sym}  \; \text{ReLU} ( \hat{A}_\text{sym}  \; \text{ReLU} (\hat{A}_\text{sym}  {X} W_0 ) \; W_1 )\; W_2 \cdots ) \; W_n ) \equiv  \text{softmax} ({Y'})
\end{equation}
For this architecture, without considering the $\text{ReLU}$ activation function, \cite{li2018deeper} shows that $Y'$ will converge to a space spanned by the eigenvectors of $\hat{A}_\text{sym}$ with eigenvalue 1. Taking activation function into consideration, our analyses on \eqref{eq1} can be summarized in the following theorems (see proof in the appendix of \cite{luan2019break}).

\begin{theorem} 1 
\label{thm1}
Suppose that $\mathcal{G}$ has $k$ connected components. Let ${X} \in \mathbb{R}^{N \times F}$ be any feature matrix and let $W_j$ be any non-negative parameter matrix with $\|W_j\|_2\leq 1$ for $j=0,1,\ldots$.
If $\mathcal{G}$ has no bipartite components, then in \eqref{eq1}, as $n \to \infty$, $\text{rank}({Y'}) \leq k$.
\end{theorem}

\begin{theorem} 2
\label{thm2}
Suppose the $n$-dimensional $\bm{x}$ and   $\bm{y}$ are independently sampled from a continuous distribution and the activation function $\text{Tanh}(z) = \frac{e^z - e^{-z}}{e^z + e^{-z}}$ is applied to $[\bm{x},\bm{y}]$ pointwisely, then
$$\doubleP(\text{rank}\left(\text{Tanh}([\bm{x},\bm{y}])\right) = \text{rank}([\bm{x},\bm{y}])) = 1$$
\end{theorem}

Theorem 1 shows that, even considering ReLU, if we simply deepen GCN as \eqref{eq1}, the extracted features will degrade under certain conditions, \ie{} ${Y'}$ only contains the stationary information of the graph structure and loses all the local information in node for being smoothed. In addition, from the proof we see that the pointwise ReLU transformation is a conspirator.  
Theorem 2 tells us that Tanh is better at keeping linear independence among column features. We design a numerical experiment on synthetic data to test, under a 100-layer GCN architecture, how activation functions affect the rank of the output in each hidden layer during the feedforward process. As Figure 1(a) shows, the rank of hidden features decreases rapidly with ReLU, while having little fluctuation under Tanh, and even the identity function performs better than ReLU. So we propose to replace ReLU by Tanh.
\begin{figure*}[htbp]
\centering
\scalebox{.85}{
\subfloat[Deep GCN]{
\captionsetup{justification = centering}
\includegraphics[width=0.32\textwidth]{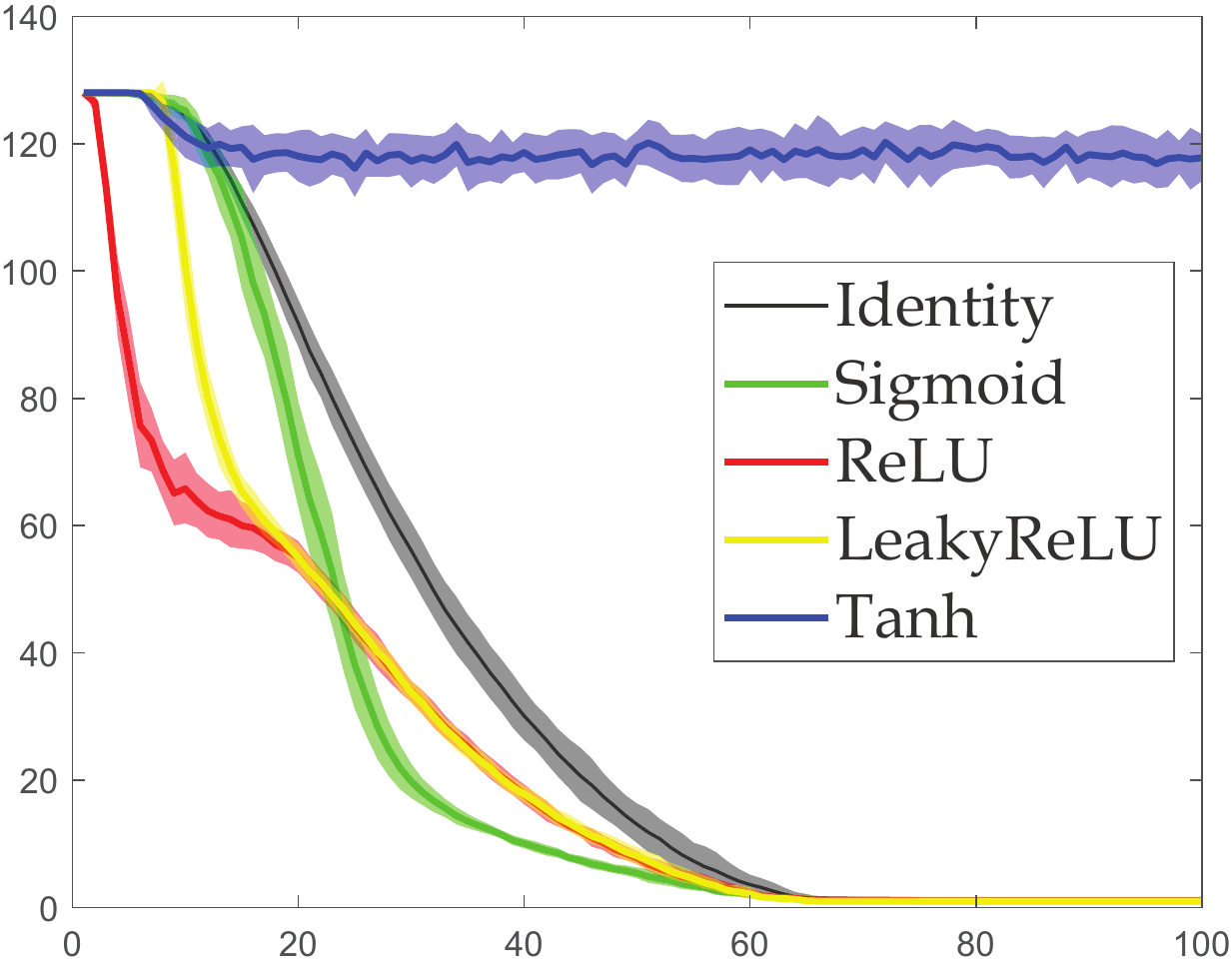}}
\hfill
\subfloat[Snowball]{
\captionsetup{justification = centering}
\includegraphics[width=0.32\textwidth]{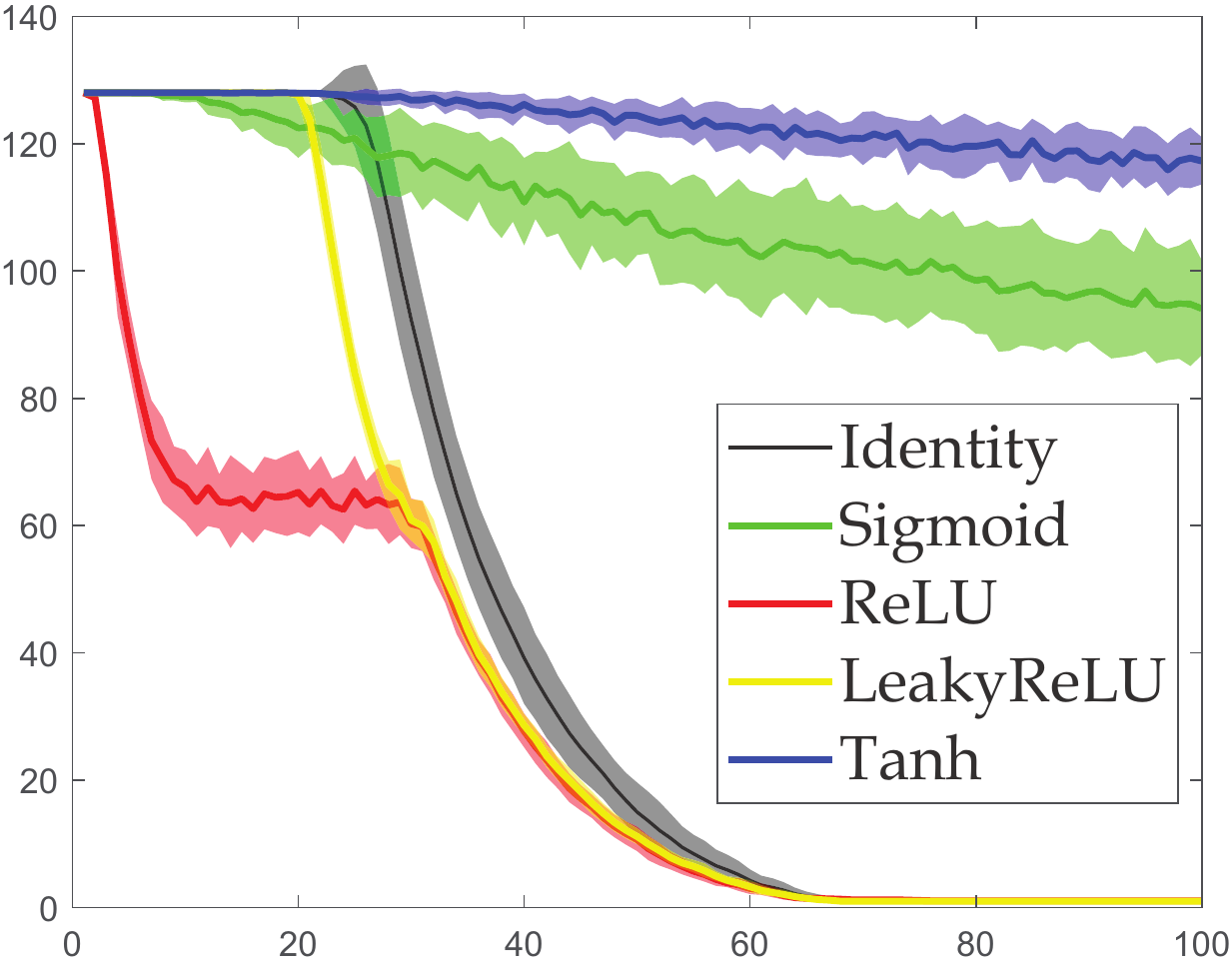}}
\hfill
\subfloat[Truncated Block Krylov]{
\captionsetup{justification = centering}
\includegraphics[width=0.32\textwidth]{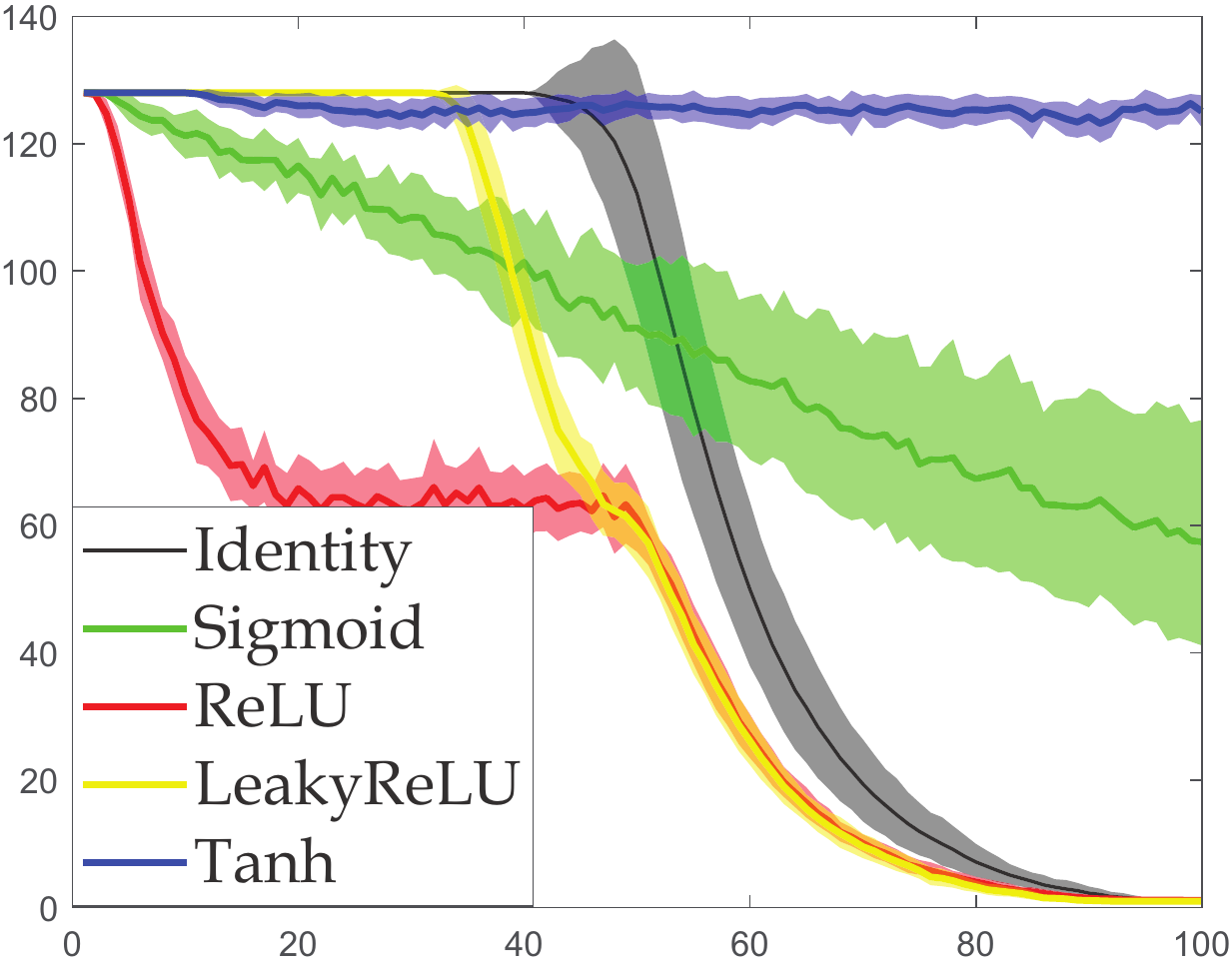}}
\caption{Changes in the number of independent features with the increment of network depth}
\label{{fig:activation}}
}
\end{figure*}

Besides activation function, to find a way to deepen GCN, we first show that any graph convolution with well-defined analytic spectral filter defined on $\hat{A}_{\text{sym}} \in \mathbb{R}^{N\times N}$ can be written as a product of a block Krylov matrix with a learnable parameter matrix in a specific form. Based on this, we propose snowball network and truncated Krylov network.
\begin{figure*}[htbp]
\centering
\scalebox{.75}{
\subfloat[Snowball]{
\captionsetup{justification = centering}
\includegraphics[width=0.49\textwidth]{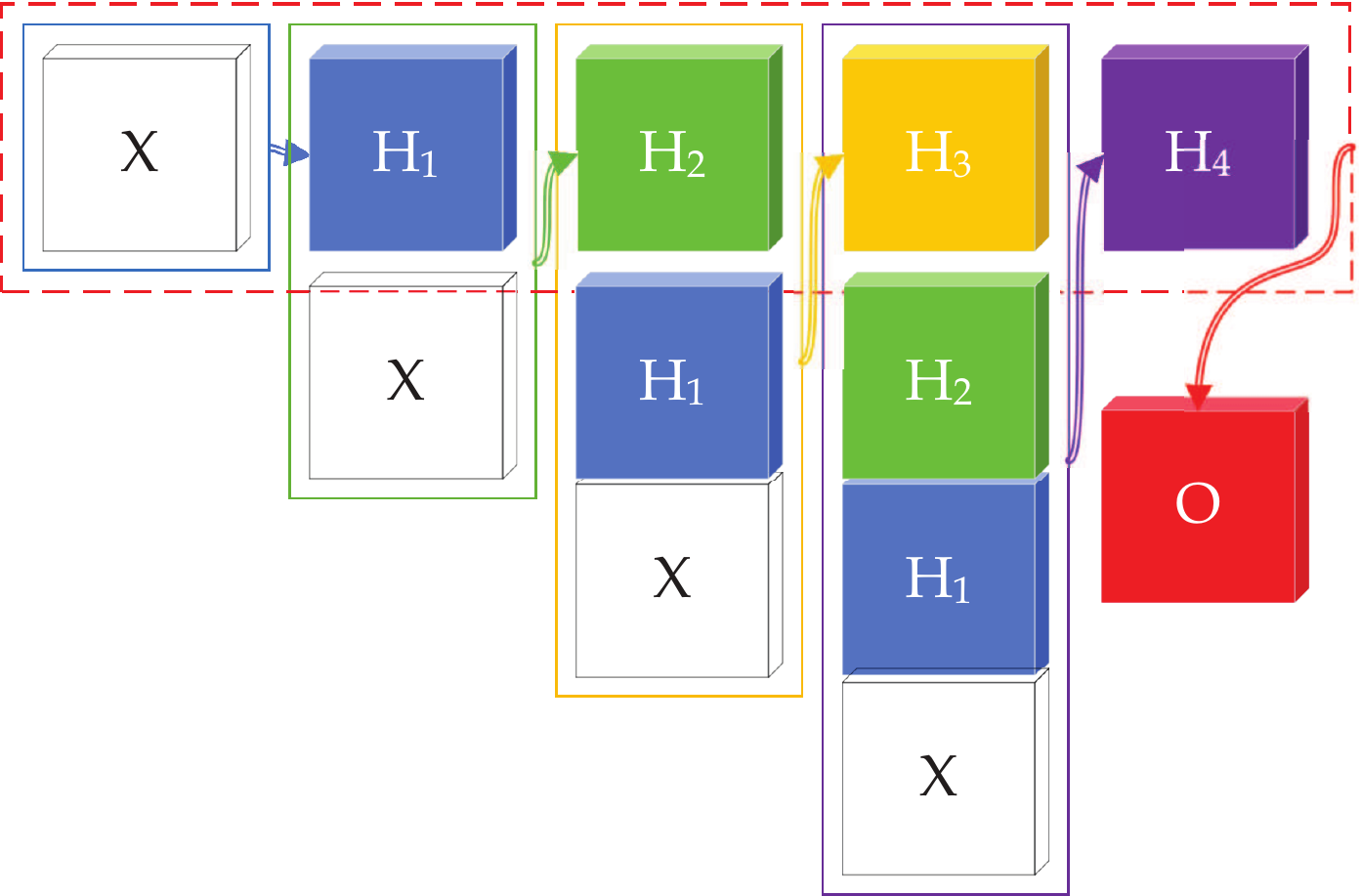}}
\hfill
\subfloat[Truncated Block Krylov]{
\captionsetup{justification = centering}
\includegraphics[width=0.49\textwidth]{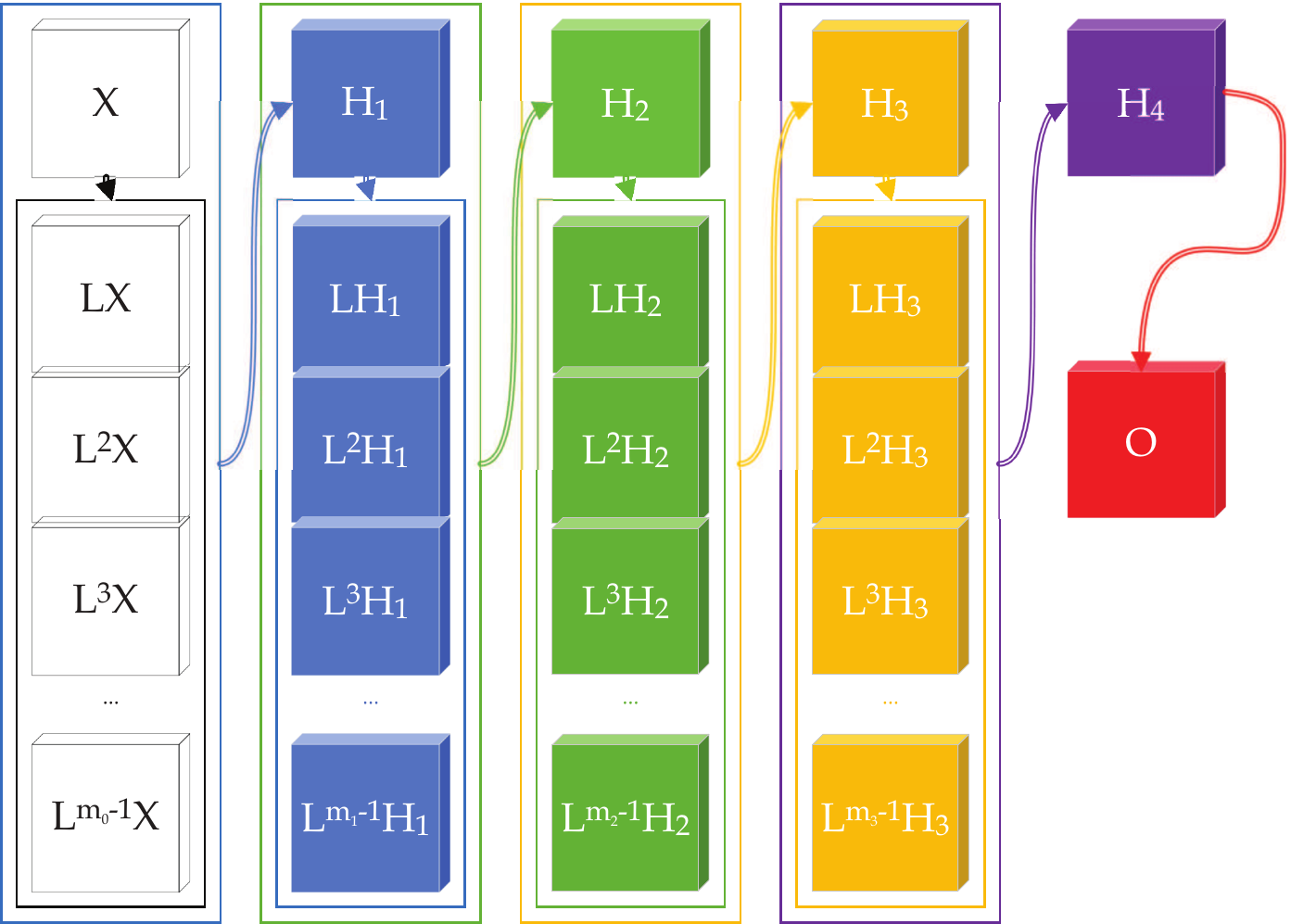}}
\caption{Snowball and Truncated Krylov Architectures}
\label{fig:architectures}
}
\end{figure*}

We take $\mathbb{S} = \mathbb{R}^{F\times F}$. Given a set of block vectors $\{X_k \}_{k=1}^m \subset \mathbb{R}^{N\times F} $, the $\mathbb{S}$-span of $\{X_k \}_{k=1}^m$ is defined as $\mathrm{span}^{\mathbb{S}} \{X_1, \dots, X_m\} \vcentcolon = \{ \sum\limits_{k=1}^{m} X_k C_k: C_k \in  \mathbb{S} \}$. Then, the order-$m$ block Krylov subspace with respect to the matrix $A \in \mathbb{R}^{N\times N}$, the block vector ${B}\in \mathbb{R}^{N\times F}$ and the vector space $\mathbb{S}$, and its corresponding block Krylov matrix are respectively defined as 
$$
\mathcal{K}_m^{\mathbb{S}} (A,{B}) \equiv \text{span}^{\mathbb{S}} \{{B}, A{B}, \dots, A^{m-1} {B}\} ,\ \ 
K_m (A, {B}) \equiv [ {B}, A {B}, \dots, A^{m-1} {B}] \in \mathbb{R}^{N\times mF}.
$$
It is shown in \cite{gutknecht2009block,frommer2017block} that there exists a smallest $m$ such that for any $k\geq m$, $A^k {B} \in \mathcal{K}_m^{\mathbb{S}} (A, {B})$, where $m$ depends on $A$ and $B$.

Let $\rho(\hat{A}_\text{sym} )$ denote the spectrum radius of $\hat{A}_\text{sym} $ and suppose $\rho(\hat{A}_\text{sym} ) < R$ where $R$ is the radius of convergence for a real analytic scalar function $g$. Based on the above definitions and conclusions, the graph convolution can be written as
\begin{equation}\label{eq5}
g(\hat{A}_\text{sym} ) X 
= \sum\limits_{n=0}^\infty \frac{g^{(n) } (0)}{n!} \hat{A}_\text{sym} ^n X 
\equiv \left[X, \hat{A}_\text{sym} X, \dots,\hat{A}_\text{sym} ^{m-1}X  \right] \left[ ({\Gamma_0}^{\mathbb{S}})^T  , ({\Gamma_1}^{\mathbb{S}})^T, \cdots ,({\Gamma_{m-1}^{\mathbb{S}}})^T  \right]^T \equiv  K_m (\hat{A}_\text{sym} ,X) \Gamma^{\mathbb{S}}
\end{equation}
where ${\Gamma_{i}^{\mathbb{S}}}\in \mathbb{R}^{F\times F}$ for $i=1,\ldots, m-1$ are parameter matrix blocks and ${\Gamma^{\mathbb{S}}}\in \mathbb{R}^{mF\times F}$. Then, a graph convolutional layer can generally be written as 
\begin{equation} \label{eq6}
g(\hat{A}_\text{sym})XW' =  K_m (\hat{A}_\text{sym},X) \Gamma^{\mathbb{S}} W' =  K_m (\hat{A}_\text{sym} ,X) W^{\mathbb{S}}
\end{equation}
where  $W' \in \mathbb{R}^{F \times O}$ is a parameter matrix, and $W^{\mathbb{S}} \equiv \Gamma^{\mathbb{S}} W' \in \mathbb{R}^{mF \times O}$. The essential  number of learnable parameters is $mF\times O$. 

The block Krylov form provides an insight about  why an architecture that concatenates multi-scale features in each layer will boost the expressive power of GCN. Based on this idea, we propose 
the snowball and truncated Block Krylov architectures \cite{luan2019break} shown in Figure \ref{fig:architectures}, where we stack multi-scale information in each layer. From the performance comparison on semi-supervised node classification tasks with different label percentage in table \ref{tab:results_no_validation}, we can see that the proposed models consistently perform better than the state-of-the-art models, especially when there are less labeled nodes. See detailed experimental results in \cite{luan2019break}.
\begin{table*}[htbp]
\setlength{\tabcolsep}{1pt}
  \centering
  \caption{Accuracy without Validation}
  \scriptsize
    \begin{tabular}{c|cccccc|cccccc|cccc}
    \multirow{2}[1]{*}{Algorithms} & \multicolumn{6}{c|}{Cora}                     & \multicolumn{6}{c|}{CiteSeer}                 & \multicolumn{4}{c}{PubMed} \\
          & 0.5\% & 1\%   & 2\%   & 3\%   & 4\%   & 5\%   & 0.5\% & 1\%   & 2\%   & 3\%   & 4\%   & 5\%   & 0.03\% & 0.05\% & 0.1\% & 0.3\% \\
    \midrule
    LP    & \cellcolor[rgb]{ .996,  .918,  .514}56.4 & \cellcolor[rgb]{ .992,  .796,  .494}62.3 & \cellcolor[rgb]{ .976,  .545,  .443}65.4 & \cellcolor[rgb]{ .973,  .412,  .42}67.5 & \cellcolor[rgb]{ .973,  .412,  .42}69.0 & \cellcolor[rgb]{ .973,  .412,  .42}70.2 & \cellcolor[rgb]{ .976,  .514,  .439}34.8 & \cellcolor[rgb]{ .973,  .412,  .42}40.2 & \cellcolor[rgb]{ .973,  .412,  .42}43.6 & \cellcolor[rgb]{ .973,  .412,  .42}45.3 & \cellcolor[rgb]{ .973,  .412,  .42}46.4 & \cellcolor[rgb]{ .973,  .412,  .42}47.3 & \cellcolor[rgb]{ .804,  .867,  .51}61.4 & \cellcolor[rgb]{ .812,  .871,  .51}66.4 & \cellcolor[rgb]{ .992,  .796,  .49}65.4 & \cellcolor[rgb]{ .973,  .412,  .42}66.8 \\
    Cheby & \cellcolor[rgb]{ .973,  .412,  .42}38.0 & \cellcolor[rgb]{ .973,  .412,  .42}52.0 & \cellcolor[rgb]{ .973,  .412,  .42}62.4 & \cellcolor[rgb]{ .98,  .596,  .455}70.8 & \cellcolor[rgb]{ .984,  .678,  .471}74.1 & \cellcolor[rgb]{ .992,  .792,  .49}77.6 & \cellcolor[rgb]{ .973,  .412,  .42}31.7 & \cellcolor[rgb]{ .976,  .482,  .431}42.8 & \cellcolor[rgb]{ .988,  .773,  .486}59.9 & \cellcolor[rgb]{ .996,  .863,  .506}66.2 & \cellcolor[rgb]{ .996,  .882,  .51}68.3 & \cellcolor[rgb]{ .996,  .886,  .51}69.3 & \cellcolor[rgb]{ .973,  .412,  .42}40.4 & \cellcolor[rgb]{ .973,  .412,  .42}47.3 & \cellcolor[rgb]{ .973,  .412,  .42}51.2 & \cellcolor[rgb]{ .984,  .698,  .475}72.8 \\
    Co-training & \cellcolor[rgb]{ .996,  .922,  .518}56.6 & \cellcolor[rgb]{ .953,  .91,  .518}66.4 & \cellcolor[rgb]{ .996,  .914,  .514}73.5 & \cellcolor[rgb]{ .996,  .882,  .51}75.9 & \cellcolor[rgb]{ .965,  .914,  .518}78.9 & \cellcolor[rgb]{ .894,  .894,  .514}80.8 & \cellcolor[rgb]{ .996,  .922,  .518}47.3 & \cellcolor[rgb]{ .992,  .847,  .502}55.7 & \cellcolor[rgb]{ .992,  .82,  .498}62.1 & \cellcolor[rgb]{ .992,  .78,  .49}62.5 & \cellcolor[rgb]{ .992,  .8,  .494}64.5 & \cellcolor[rgb]{ .992,  .804,  .494}65.5 & \cellcolor[rgb]{ .757,  .855,  .506}62.2 & \cellcolor[rgb]{ .655,  .824,  .498}68.3 & \cellcolor[rgb]{ .702,  .835,  .502}72.7 & \cellcolor[rgb]{ .847,  .878,  .51}78.2 \\
    Self-training & \cellcolor[rgb]{ .992,  .843,  .502}53.7 & \cellcolor[rgb]{ .973,  .914,  .518}66.1 & \cellcolor[rgb]{ .988,  .918,  .518}73.8 & \cellcolor[rgb]{ .933,  .906,  .518}77.2 & \cellcolor[rgb]{ .894,  .894,  .514}79.4 & \cellcolor[rgb]{ .996,  .914,  .514}80.0 & \cellcolor[rgb]{ .992,  .792,  .49}43.3 & \cellcolor[rgb]{ .996,  .914,  .514}58.1 & \cellcolor[rgb]{ .694,  .835,  .502}68.2 & \cellcolor[rgb]{ .773,  .859,  .506}69.8 & \cellcolor[rgb]{ .894,  .89,  .514}70.4 & \cellcolor[rgb]{ .969,  .914,  .518}71.0 & \cellcolor[rgb]{ .988,  .745,  .482}51.9 & \cellcolor[rgb]{ .988,  .753,  .482}58.7 & \cellcolor[rgb]{ .992,  .835,  .498}66.8 & \cellcolor[rgb]{ .996,  .898,  .51}77.0 \\
    Union & \cellcolor[rgb]{ .922,  .902,  .514}58.5 & \cellcolor[rgb]{ .757,  .855,  .506}69.9 & \cellcolor[rgb]{ .784,  .859,  .506}75.9 & \cellcolor[rgb]{ .788,  .863,  .506}78.5 & \cellcolor[rgb]{ .753,  .851,  .506}80.4 & \cellcolor[rgb]{ .757,  .855,  .506}81.7 & \cellcolor[rgb]{ .996,  .89,  .51}46.3 & \cellcolor[rgb]{ .937,  .906,  .518}59.1 & \cellcolor[rgb]{ .973,  .914,  .518}66.7 & \cellcolor[rgb]{ .996,  .871,  .506}66.7 & \cellcolor[rgb]{ .996,  .871,  .506}67.6 & \cellcolor[rgb]{ .996,  .863,  .506}68.2 & \cellcolor[rgb]{ .973,  .914,  .518}58.4 & \cellcolor[rgb]{ .996,  .914,  .514}64.0 & \cellcolor[rgb]{ .922,  .898,  .514}70.7 & \cellcolor[rgb]{ .635,  .82,  .498}79.2 \\
    Intersection & \cellcolor[rgb]{ .988,  .733,  .478}49.7 & \cellcolor[rgb]{ .996,  .898,  .51}65.0 & \cellcolor[rgb]{ .996,  .886,  .51}72.9 & \cellcolor[rgb]{ .945,  .906,  .518}77.1 & \cellcolor[rgb]{ .894,  .894,  .514}79.4 & \cellcolor[rgb]{ .988,  .918,  .518}80.2 & \cellcolor[rgb]{ .992,  .78,  .49}42.9 & \cellcolor[rgb]{ .937,  .906,  .518}59.1 & \cellcolor[rgb]{ .616,  .812,  .498}68.6 & \cellcolor[rgb]{ .698,  .835,  .502}70.1 & \cellcolor[rgb]{ .796,  .863,  .506}70.8 & \cellcolor[rgb]{ .925,  .902,  .514}71.2 & \cellcolor[rgb]{ .988,  .749,  .482}52.0 & \cellcolor[rgb]{ .988,  .773,  .486}59.3 & \cellcolor[rgb]{ .996,  .914,  .514}69.7 & \cellcolor[rgb]{ .973,  .914,  .518}77.6 \\
    MultiStage & \cellcolor[rgb]{ .82,  .871,  .51}61.1 & \cellcolor[rgb]{ .996,  .851,  .502}63.7 & \cellcolor[rgb]{ .929,  .902,  .514}74.4 & \cellcolor[rgb]{ .996,  .89,  .51}76.1 & \cellcolor[rgb]{ .992,  .843,  .502}77.2 &       & \cellcolor[rgb]{ .722,  .843,  .502}53.0 & \cellcolor[rgb]{ .996,  .906,  .514}57.8 & \cellcolor[rgb]{ .996,  .859,  .506}63.8 & \cellcolor[rgb]{ .996,  .902,  .514}68.0 & \cellcolor[rgb]{ .996,  .898,  .51}69.0 &       & \cellcolor[rgb]{ .996,  .906,  .514}57.4 & \cellcolor[rgb]{ .988,  .922,  .518}64.3 & \cellcolor[rgb]{ .976,  .914,  .518}70.2 &  \\
    M3S   & \cellcolor[rgb]{ .804,  .867,  .51}61.5 & \cellcolor[rgb]{ .91,  .898,  .514}67.2 & \cellcolor[rgb]{ .812,  .871,  .51}75.6 & \cellcolor[rgb]{ .867,  .886,  .514}77.8 & \cellcolor[rgb]{ .996,  .886,  .51}78.0 &       & \cellcolor[rgb]{ .573,  .8,  .494}56.1 & \cellcolor[rgb]{ .702,  .839,  .502}62.1 & \cellcolor[rgb]{ .996,  .918,  .514}66.4 & \cellcolor[rgb]{ .647,  .82,  .498}70.3 & \cellcolor[rgb]{ .871,  .886,  .514}70.5 &       & \cellcolor[rgb]{ .929,  .902,  .514}59.2 & \cellcolor[rgb]{ .98,  .918,  .518}64.4 & \cellcolor[rgb]{ .929,  .902,  .514}70.6 &  \\
    GCN   & \cellcolor[rgb]{ .976,  .537,  .443}42.6 & \cellcolor[rgb]{ .98,  .596,  .455}56.9 & \cellcolor[rgb]{ .984,  .655,  .463}67.8 & \cellcolor[rgb]{ .992,  .824,  .498}74.9 & \cellcolor[rgb]{ .996,  .863,  .506}77.6 & \cellcolor[rgb]{ .996,  .878,  .506}79.3 & \cellcolor[rgb]{ .973,  .467,  .427}33.4 & \cellcolor[rgb]{ .98,  .588,  .451}46.5 & \cellcolor[rgb]{ .992,  .831,  .498}62.6 & \cellcolor[rgb]{ .996,  .875,  .506}66.9 & \cellcolor[rgb]{ .996,  .894,  .51}68.7 & \cellcolor[rgb]{ .996,  .894,  .51}69.6 & \cellcolor[rgb]{ .98,  .584,  .451}46.4 & \cellcolor[rgb]{ .973,  .482,  .431}49.7 & \cellcolor[rgb]{ .976,  .549,  .443}56.3 & \cellcolor[rgb]{ .996,  .878,  .51}76.6 \\
    GCN-SVAT & \cellcolor[rgb]{ .98,  .565,  .447}43.6 & \cellcolor[rgb]{ .973,  .482,  .431}53.9 & \cellcolor[rgb]{ .992,  .82,  .498}71.4 & \cellcolor[rgb]{ .996,  .863,  .506}75.6 & \cellcolor[rgb]{ .996,  .902,  .514}78.3 & \cellcolor[rgb]{ .992,  .835,  .498}78.5 & \cellcolor[rgb]{ .996,  .914,  .514}47.0 & \cellcolor[rgb]{ .988,  .753,  .482}52.4 & \cellcolor[rgb]{ .996,  .902,  .514}65.8 & \cellcolor[rgb]{ .996,  .914,  .514}68.6 & \cellcolor[rgb]{ .996,  .91,  .514}69.5 & \cellcolor[rgb]{ .996,  .918,  .514}70.7 & \cellcolor[rgb]{ .988,  .749,  .482}52.1 & \cellcolor[rgb]{ .984,  .702,  .475}56.9 & \cellcolor[rgb]{ .988,  .745,  .482}63.5 & \cellcolor[rgb]{ .996,  .906,  .514}77.2 \\
    GCN-DVAT & \cellcolor[rgb]{ .988,  .714,  .475}49 & \cellcolor[rgb]{ .992,  .78,  .49}61.8 & \cellcolor[rgb]{ .992,  .839,  .502}71.9 & \cellcolor[rgb]{ .996,  .882,  .51}75.9 & \cellcolor[rgb]{ .996,  .906,  .514}78.4 & \cellcolor[rgb]{ .992,  .843,  .502}78.6 & \cellcolor[rgb]{ .792,  .863,  .506}51.5 & \cellcolor[rgb]{ .984,  .918,  .518}58.5 & \cellcolor[rgb]{ .843,  .878,  .51}67.4 & \cellcolor[rgb]{ .925,  .902,  .514}69.2 & \cellcolor[rgb]{ .796,  .863,  .506}70.8 & \cellcolor[rgb]{ .906,  .894,  .514}71.3 & \cellcolor[rgb]{ .992,  .784,  .49}53.3 & \cellcolor[rgb]{ .988,  .753,  .482}58.6 & \cellcolor[rgb]{ .992,  .82,  .498}66.3 & \cellcolor[rgb]{ .996,  .914,  .514}77.3 \\
    \midrule
    \textit{\textbf{linear Snowball}} & \cellcolor[rgb]{ .561,  .796,  .494}\textit{\textbf{67.6}} & \cellcolor[rgb]{ .498,  .776,  .49}\textit{\textbf{74.6}} & \cellcolor[rgb]{ .49,  .776,  .49}\textit{\textbf{78.9}} & \cellcolor[rgb]{ .522,  .784,  .49}\textit{\textbf{80.9}} & \cellcolor[rgb]{ .482,  .773,  .49}\textit{\textbf{82.3}} & \cellcolor[rgb]{ .569,  .8,  .494}\textit{\textbf{82.9}} & \cellcolor[rgb]{ .576,  .8,  .494}\textit{\textbf{56.0}} & \cellcolor[rgb]{ .6,  .808,  .498}\textit{\textbf{63.4}} & \cellcolor[rgb]{ .494,  .776,  .49}\textit{\textbf{69.3}} & \cellcolor[rgb]{ .561,  .796,  .494}\textit{\textbf{70.6}} & \cellcolor[rgb]{ .388,  .745,  .482}\textit{\textbf{72.5}} & \cellcolor[rgb]{ .62,  .812,  .498}\textit{\textbf{72.6}} & \cellcolor[rgb]{ .573,  .8,  .494}\textit{\textbf{65.5}} & \cellcolor[rgb]{ .635,  .82,  .498}\textit{\textbf{68.5}} & \cellcolor[rgb]{ .604,  .808,  .498}\textit{\textbf{73.6}} & \cellcolor[rgb]{ .533,  .788,  .494}\textit{\textbf{79.7}} \\
    \textit{\textbf{Snowball}} & \cellcolor[rgb]{ .525,  .784,  .49}\textit{\textbf{68.4}} & \cellcolor[rgb]{ .576,  .8,  .494}\textit{\textbf{73.2}} & \cellcolor[rgb]{ .541,  .792,  .494}\textit{\textbf{78.4}} & \cellcolor[rgb]{ .525,  .788,  .494}\textit{\textbf{80.8}} & \cellcolor[rgb]{ .482,  .773,  .49}\textit{\textbf{82.3}} & \cellcolor[rgb]{ .557,  .796,  .494}\textit{\textbf{83.0}} & \cellcolor[rgb]{ .553,  .796,  .494}\textit{\textbf{56.4}} & \cellcolor[rgb]{ .565,  .796,  .494}\textit{\textbf{63.9}} & \cellcolor[rgb]{ .592,  .804,  .494}\textit{\textbf{68.7}} & \cellcolor[rgb]{ .584,  .804,  .494}\textit{\textbf{70.5}} & \cellcolor[rgb]{ .553,  .792,  .494}\textit{\textbf{71.8}} & \cellcolor[rgb]{ .584,  .804,  .494}\textit{\textbf{72.8}} & \cellcolor[rgb]{ .514,  .784,  .49}\textit{\textbf{66.5}} & \cellcolor[rgb]{ .631,  .816,  .498}\textit{\textbf{68.6}} & \cellcolor[rgb]{ .647,  .82,  .498}\textit{\textbf{73.2}} & \cellcolor[rgb]{ .439,  .761,  .486}\textit{\textbf{80.1}} \\
    \textit{\textbf{truncated Krylov}} & \cellcolor[rgb]{ .388,  .745,  .482}\textit{\textbf{71.8}} & \cellcolor[rgb]{ .388,  .745,  .482}\textit{\textbf{76.5}} & \cellcolor[rgb]{ .388,  .745,  .482}\textit{\textbf{80.0}} & \cellcolor[rgb]{ .388,  .745,  .482}\textit{\textbf{82.0}} & \cellcolor[rgb]{ .388,  .745,  .482}\textit{\textbf{83.0}} & \cellcolor[rgb]{ .388,  .745,  .482}\textit{\textbf{84.1}} & \cellcolor[rgb]{ .388,  .745,  .482}\textit{\textbf{59.9}} & \cellcolor[rgb]{ .388,  .745,  .482}\textit{\textbf{66.1}} & \cellcolor[rgb]{ .388,  .745,  .482}\textit{\textbf{69.8}} & \cellcolor[rgb]{ .388,  .745,  .482}\textit{\textbf{71.3}} & \cellcolor[rgb]{ .424,  .757,  .486}\textit{\textbf{72.3}} & \cellcolor[rgb]{ .388,  .745,  .482}\textit{\textbf{73.7}} & \cellcolor[rgb]{ .388,  .745,  .482}\textit{\textbf{68.7}} & \cellcolor[rgb]{ .388,  .745,  .482}\textit{\textbf{71.4}} & \cellcolor[rgb]{ .388,  .745,  .482}\textit{\textbf{75.5}} & \cellcolor[rgb]{ .388,  .745,  .482}\textit{\textbf{80.4}} \\
    \bottomrule
    \bottomrule
    \multicolumn{17}{m{0.75\textwidth}}{\scriptsize For each (column), the greener the cell, the better the performance. The redder, the worse. If our methods achieve better performance than all others, the corresponding cell will be in bold.}
    \end{tabular}%
\label{tab:results_no_validation}%
\end{table*}%

\subsection{Future Works on Over-smoothing}
\label{sec:remaining_oversmoothing}
\paragraph{Weight Initialization for GNNs} Even without aggregation in each hidden layer, an NN with deep architecture still suffers from vanishing activation variances and back-propagated gradients variance  problem \cite{glorot2010understanding}, which make the training of deep NN hard. In last decade, designing new parameter initialization methods is proved to be effective
 \cite{glorot2010understanding, he2015delving} to address the variance reduction problem during feedforward and backpropagation process. This motivates us to investigate the variance propagation in GNNs and analyze if the current weight initialization methods are suitable for GNNs or not. To this end, we can show that the vanishing variance caused by aggregation operation in GNNs is more serious than NN. Designing a new parameter initialization scheme for GNNs is potentially a feasible way to address this problem and empirically achieves promising performance \cite{luan2020training}. we will propose a new method in this subsection.

The current initialization scheme of GNNs still follows the Xavier initialization \cite{glorot2010understanding}, \ie{} $W_i \sim U\left[-\frac{\sqrt{6}}{\sqrt{n_{j}+n_{j+1}}}, \frac{\sqrt{6}}{\sqrt{n_{j}+n_{j+1}}}\right]$, or He (or Kaiming) initialization \cite{he2015delving}, \ie{} $W_i \sim N\left(0, \sqrt{2/n_i} \right)$, which is designed for traditional multilayer perceptron (MLP) , where $W_i$ is the parameter matrix of layer $i$ and $n_i$ is the number of hidden units of layer $i$. These two initialization methods are derived by studying the variance propagation between layers during feedforward and backpropagation process. These two processes are different in GNNs by an extra multiplication of aggregation operator $\hat{A}$. To analyze the variance propagation, we use deep GCN as an example, use $\hat{A} = \hat{A}_\text{rw}$ and decompose it as follows,
\begin{equation}\label{eq:deep_gcn_decompose}
\begin{aligned}
   & Y_0 = X, \;H_1 = \hat{A}_\text{rw} X W_0, \; Y_1 = f (H_1),\ H_{l+1} = \hat{A}_\text{rw} Y_l W_l, \; Y_{l+1} = f (H_{l+1}), \; l = 1, \dots, n\\
   & Y = \text{softmax} (\hat{A}_\text{rw}  {Y_n} W_n ) \equiv  \text{softmax} (H_{n+1}), \ \mathcal{L} = -\trace ({Z}^T \text{log} Y)\\
\end{aligned}
\end{equation}
where $H_l,Y_l \in \mathbb{R}^{N \times F_{l}}$, $W_l \in \mathbb{R}^{F_l \times F_{l+1}}$; ${Z}\in\Rbb^{N\times C}$ is the ground truth matrix with one-hot label vector. Then the gradient propagates in the following way,
\begin{equation}\label{eq:gradient}
\begin{aligned}
    \;& \frac{\partial \mathcal{L}}{\partial H_l} = \frac{\partial \mathcal{L} }{\partial Y_l} \odot f'(H_l), \;
   \frac{\partial \mathcal{L}}{\partial W_{l-1}} = Y_{l-1}^T \hat{A}_\text{rw} \frac{\partial \mathcal{L} }{\partial H_l }, \ \frac{\partial \mathcal{L}}{\partial Y_{l-1}} = \hat{A}_\text{rw} \frac{\partial \mathcal{L}}{ \partial H_l}  W_{l-1}^T  \\
\end{aligned}
\end{equation}

\paragraph{Variance Analysis: Forward View}
\label{sec:variance_forward_view}
Consider element $i,j$ in matrix $H_{l+1}$ during the feed-forward process in \eqref{eq:deep_gcn_decompose},
\begin{equation}
\label{eq:feedforward_nodewise_propagation}
\left(H_{l+1}\right)_{ij} = (\hat{A}_\text{rw})_{i,:} Y_l (W_l)_{:,j} = \sum\limits_{t=1}^{F_l} \sum\limits_{k=1}^N (\hat{A}_\text{rw})_{ik} \left(Y_l\right)_{kt} (W_l)_{t,j}, \ Y_{l+1} = f (H_{l+1}), \; l = 1, \dots, n 
\end{equation}
Suppose we have linear activation function such as that proposed in \cite{wu2019simplifying}; each element in $W_l$ is \iid initialized with $E\left( (W_l)_{ij} \right) = 0$; $E\left( (Y_l)_{kt} \right) = 0$ and all elements in $Y_l$ are independent  \footnote{For simplicity, the independence assumption is directly borrowed from \cite{glorot2010understanding}, but theoretically it is too strong for GNNs. We will try to relax this assumption in the future.}.
Then, $\text{Var}\left((Y_{l+1})_{ij}\right) = \text{Var}\left((Y_{l+1})\right)$  can be written as

\begin{equation}
\begin{aligned} \label{eq:variance_hidden_units}
   & \text{Var}\left(\sum\limits_{t=1}^{F_l} \sum\limits_{k=1}^N (\hat{A}_\text{rw})_{ik} \left(Y_l\right)_{kt} (W_l)_{t,j}\right) = \sum\limits_{t=1}^{F_l} \sum\limits_{k=1}^N \text{Var}\left( (\hat{A}_\text{rw})_{ik} \left(Y_l\right)_{kt} (W_l)_{t,j}\right)
    = \frac{F_l}{d_i+1} \text{Var}\left(Y_l) \text{Var}(W_l\right) 
\end{aligned}
\end{equation}

Suppose each element in $Y_{l}$ shares the same variance denoted as $\text{Var}\left(Y_{l}\right)$. To prevent variance vanishing between layers, \ie{} $\text{Var}\left(Y_{l+1}\right) = \text{Var} \left({Y_{l}}\right)$, from \eqref{eq:feedforward_nodewise_propagation} we can approximately have (see computation in Appendix \ref{appendix:forward_view})

\begin{equation}
\label{eq:feedforward_weight_initialization}
    \text{Var}(W_l) = \frac{d_i+1}{F_l}
\end{equation}
This tells us that the variance of $W_l$ depends on the degree of a node, but since the parameter matrix is shared by all nodes, we cannot design a node specified initialization scheme. Thus, we make a compromise between nodes as follows

\begin{equation}
\label{eq:variance_forwad_average}
    \text{Var}(W_l) \approx \frac{\sum\limits_{i=1}^N (d_i+1)}{N F_l} = \frac{1+\text{average node degree}}{F_l}
\end{equation}

Another way is to use weighted average by considering the node degree as the weight of each node. Through this way, we have
\begin{equation}
\label{eq:variance_forwad_weighted_average}
    \text{Var}(W) = \sum\limits_{i=1}^N \frac{d_i+1}{\sum\limits_{j=1}^N d_j+1}\frac{d_i+1}{F_l} = \frac{\sum\limits_{i=1}^N (d_i+1)^2}{ (\sum\limits_{i=1}^N d_i+1) F_l}
\end{equation}

\paragraph{Variance Analysis: Backward View}
\label{sec:variance_backward_view}
Under the same assumption as feedforward view and suppose each element in $\frac{\partial \mathcal{L}}{\partial H_l}$ and $\frac{\partial \mathcal{L}}{\partial W_{l-1}}$ are independent to each other and has zero mean, from \eqref{eq:gradient} we can approximately (see computation in Appendix \ref{appendix:backward_view})

\begin{equation}\label{eq:backward_view}
\begin{aligned}
  \frac{\partial \mathcal{L}}{\partial H_l} = \frac{\partial \mathcal{L}}{\partial Y_l} = \hat{A}_\text{rw} \frac{\partial \mathcal{L}}{ \partial H_{l+1}}  W_{l}^T, \;
   \frac{\partial \mathcal{L}}{\partial W_{l-1}} = Y_{l-1}^T \hat{A}_\text{rw} \frac{\partial \mathcal{L} }{\partial H_l }\\
\end{aligned}
\end{equation}
Then,
\begin{equation}
\begin{aligned}
   \left(\frac{\partial \mathcal{L}}{\partial H_l}\right)_{ij} & = \sum\limits_{t=1}^{F_{l+1}} \sum\limits_{k=1}^N (\hat{A}_\text{rw})_{ik} \left(\frac{\partial \mathcal{L}}{\partial H_{l+1}}\right)_{kt} (W_l^T)_{t,j} \\
   \left(\frac{\partial \mathcal{L}}{\partial W_{l-1}}\right)_{ij} & = (\hat{A}_\text{rw} Y_{l-1})_{\cdot i}^T  \left(\frac{\partial \mathcal{L}}{\partial H_l }\right)_{\cdot j} = \sum\limits_{k=1}^N  (\sum\limits_{t=1}^N (\hat{A}_\text{rw})_{kt} (Y_{l-1})_{t i}) \left(\frac{\partial \mathcal{L}}{\partial H_l }\right)_{k j}, \\
   \end{aligned}
\end{equation}
Thus,
\begin{equation}
\begin{aligned}
\label{eq:backward_variance}
   \text{Var}\left(\left(\frac{\partial \mathcal{L}}{\partial H_l}\right)_{ij}\right) & = \text{Var}\left( \sum\limits_{t=1}^{F_{l+1}} \sum\limits_{k=1}^N \hat{A}_{\text{rw}_{ik}} \left(\frac{\partial \mathcal{L}}{\partial H_{l+1}}\right)_{kt} (W_l^T)_{t,j} \right)
    = \frac{F_{l+1}}{d_i+1} \text{Var}\left( \frac{\partial \mathcal{L}}{\partial H_{l+1}}  \right) \text{Var}\left(W_l\right) \\
    \text{Var}\left(\left(\frac{\partial \mathcal{L}}{\partial W_{l-1}}\right)_{ij}\right) & = \text{Var}\left( \sum\limits_{k=1}^N  (\sum\limits_{t=1}^N (\hat{A}_\text{rw})_{kt} (Y_{l-1})_{t i}) \left(\frac{\partial \mathcal{L}}{\partial H_l }\right)_{k j} \right) 
     = \left(\sum\limits_{k=1}^N \frac{1}{d_k+1} \right)\text{Var}\left(Y_{l-1} \right) \text{Var} \left(\frac{\partial \mathcal{L}}{\partial H_l } \right) \\
   \end{aligned}
\end{equation}

\begin{equation}
    \text{Var}(W_l) = \frac{\sum\limits_{i=1}^N (d_i+1)}{N F_{l+1}} \approx \frac{1+\text{average node degree}}{F_{l+1}} 
\end{equation}

From \eqref{eq:variance_hidden_units}\eqref{eq:backward_variance},  $\text{Var}\left(Y_{l-1} \right)$ can be approximately written as
\begin{equation}
\begin{aligned}
\label{eq:variance_flatten}
   & \text{Var} \left(\frac{\partial \mathcal{L}}{\partial H_l}\right) \approx \frac{N F_{l+1}}{\sum\limits_{i=1}^N d_i+1} \text{Var}\left( \frac{\partial \mathcal{L}}{\partial H_{l+1}}  \right) \text{Var}\left(W_l\right) \approx \text{Var}\left(\frac{\partial \mathcal{L}}{\partial H_{n+1}}\right) \prod\limits_{l'=l+1}^{n+1} \frac{N F_{l'}}{\sum\limits_{i=1}^N d_i+1} \text{Var}\left(W_{l'-1}\right) \\
   & \text{Var}\left(Y_{l-1}\right) \approx \frac{N F_{l-2}}{\sum_k d_k+1} \text{Var}\left(\bm{Y_{l-2}}) \text{Var}(W_{l-2}\right) \approx \text{Var}(\bm{Y_{0}}) \prod\limits_{l'=0}^{l-2}\frac{N F_{l'}}{\sum_k d_k+1}  \text{Var}(W_{l'})\\
\end{aligned}
\end{equation}
From \eqref{eq:backward_variance}, if each $\text{Var}(W_{l'})$ equals to $\text{Var}(W)$ and each $F_l$ equals to $F$, then
\begin{equation}
\begin{aligned}
\label{eq:gradient_parameter_matrix}
    & \text{Var}\left(\left(\frac{\partial \mathcal{L}}{\partial W_{l-1}}\right)_{ij}\right) \approx \left(\sum\limits_{k=1}^N \frac{1}{d_k+1} \right)\text{Var}(\bm{Y_{0}}) \text{Var}\left(\frac{\partial \mathcal{L}}{\partial H_{n+1}}  \right) \left(\frac{N F}{\sum_k d_k+1}\text{Var}(W) \right)^n  \\
   \end{aligned}
\end{equation}

Combined with \eqref{eq:variance_forwad_average}, we can set the variance of the parameter matrix as
\begin{equation}
\label{eq:weight_initialization_variance}
     \text{Var}(W_l) \approx \frac{2\sum\limits_{i=1}^N (d_i+1)}{N (F_l+F_{l+1})} = \frac{2(1+\text{average node degree})}{(F_l + F_{l+1})}
\end{equation}
Thus, each element in $W_i$ can be drawn from $N(0, \sqrt{\frac{2(1+\text{average node degree})}{(F_l + F_{l+1})}})$.

\paragraph{Adaptive ReLU (AdaReLU) Activation Function}
To satisfy the assumption that the activation function is linear at the beginning of training process and to still learn a nonlinear function during training, we design the following adaptive ReLU (AdaReLU) activation function
$$f(\bm{x}_{i}) = \left\{
\begin{array}{ll}
\beta_i \bm{x}_{i}, & \text { if } \bm{x}_{i}>0 \\ 
\alpha_{i} \bm{x}_{i}, & \text { if } \bm{x}_{i} \leq 0
\end{array}
\right.$$
where $\alpha_i$ and $\beta_i$ are learnable parameters and are initialized to be 1. If $\alpha_i = \alpha$ and $\beta_i= \beta$ for all $i$, we have channel-shared  AdaReLU, otherwise we have channel-wise AdaReLU \footnote{The words "channel-shared" and "channel-wise" are borrowed from \cite{he2015delving}, which indicate if we share the same learning parameter between each feature dimension or not.}. From the preliminary experimental results, channel-wise works better than channel-shared AdaReLU. 

There exist some experimental evidence \cite{luan2020training} that controlling variance flow by initialization like \eqref{eq:weight_initialization_variance} can relieve the performance decrease of deep GCN. But more tests and hyperparameter tunning still needs to be done. More theoretical analysis on variance propagation needs to be done.

\section{GNNs on Heterophily Graphs}
\label{sec:heterophily}
GNNs are considered as an extension of basic Neural Networks (NNs) by additionally making use of graph structure based on the relational inductive bias (homophily assumption), rather than treating the nodes as collections of independent and identically distributed (\iid) samples. Though GNNs are believed to outperform basic NNs in real-world tasks, it is found that in some cases, the graph-aware models have little performance gain or even underperform graph-agnostic models \cite{pei2020geom, zhu2020generalizing,luan2020complete,  chien2021adaptive, zhu2020graph}. One of the main reasons for the performance degradation is believed to be heterophily, \ie{} when the connected nodes tend to have different labels \cite{zhu2020generalizing,zhu2020graph}. Heterophily challenge has received attention recently and there are increasing number of models being put forward to analyze \cite{luan2022we,luan2023graph} and address this problem  \cite{zhu2020beyond, liu2020non, luan2021heterophily, chien2021adaptive,zhu2020graph,yan2021two,luan2022revisiting}. 

In this section , we  first introduce the most commently used homophily metrics in subsection \ref{sec:homophily_metrics}. Then, we  show that not all cases of heterophily are harmful for GNNs and propose new metrics based on a similarity matrix which considers the influence of both graph structure and input features on GNNs in subsection \ref{sec:heterophily_analysis}. The metrics demonstrate advantages over the commonly used homophily metrics by tests on synthetic graphs. From the metrics and the observations, we find some cases of harmful heterophily can be addressed by diversification operation and its effectiveness can be proved in subsection \ref{sec:diversification_distinguishability}. With this fact and knowledge of filterbanks, we propose the Adaptive Channel Mixing (ACM) framework  in subsection \ref{sec:acm_gnn_architecture} to adaptively exploit aggregation, diversification and identity channels in each GNN layer to address harmful heterophily. We validate the ACM-augmented baselines with real-world node classification tasks. They consistently achieve significant performance gain and exceed the state-of-the-art GNNs on most of the tasks without incurring significant computational burden. In subsection \ref{sec:heterophily_prior_work}, we introduce some prior work on addressing heterophily problems and explain their differences with ACM framework. The limitation of diversification operation and remaining challenges of heterophily problems are discussed in subsection \ref{sec:adj_learner_deeper_thinking}

\subsection{Metrics of Homophily}
\label{sec:homophily_metrics}
The metrics of homophily are defined by considering different relations between node labels and graph structures defined by adjacency matrix. There are three commonly used homophily metrics: edge homophily \cite{abu2019mixhop,zhu2020beyond}, node homophily \cite{pei2020geom}, and class homophily \cite{lim2021new} \footnote{The authors in \cite{lim2021new} did not name this homophily metric. We name it class homophily based on its definition.} defined as follows:
\begin{equation}
\begin{aligned}
\label{eq:definition_homophily_metrics}
    &H_\text{edge}(\mathcal{G}) = \frac{\big|\{e_{uv} \mid e_{uv}\in \mathcal{E}, Z_{u,:}=Z_{v,:}\}\big|}{|\mathcal{E}|}, \ \ 
    H_\text{node}(\mathcal{G}) = \frac{1}{|\mathcal{V}|} \sum_{v \in \mathcal{V}} 
    \frac{\big|\{u \mid u \in \mathcal{N}_v, Z_{u,:}=Z_{v,:}\}\big|}{d_v}, \\
    &H_\text{class}(\mathcal{G}) = \frac{1}{C-1} \sum_{k=1}^{C}\left[h_{k}
    -\frac{\big|\{v \mid Z_{v,k} = 1 \}\big|}{N}\right]_{+}, \ \ 
    h_{k}=\frac{\sum_{v \in \mathcal{V}} \big|\{u \mid Z_{v,k} = 1, u \in \mathcal{N}_v,   Z_{u,:}=Z_{v,:}\}\big| }{\sum_{v \in \{v|Z_{v,k}=1\}} d_{v}}
\end{aligned}
\end{equation}
where $[a]_{+}=\max (a, 0)$; $h_{k}$ is the class-wise homophily metric \cite{lim2021new}. They are all in the range of $[0,1]$ and a value close to $1$ corresponds to strong homophily while a value close to $0$ indicates strong heterophily. $H_\text{edge}(\mathcal{G})$ measures the proportion of edges that connect two nodes in the same class; $H_\text{node}(\mathcal{G})$ evaluates the average proportion of edge-label consistency of all nodes; $H_\text{class}(\mathcal{G})$ tries to avoid the sensitivity to imbalanced class, which can cause $H_\text{edge}$ misleadingly large. The above definitions are all based on the graph-label consistency and imply that the inconsistency will cause harmful effect to the performance of GNNs. With this in mind, we will show a counter example to illustrate the insufficiency of the above metrics and propose new metrics in the following subsection.

\begin{wrapfigure}{R}{0.5\textwidth}
  \begin{center}
    \includegraphics[width=1.0\textwidth]{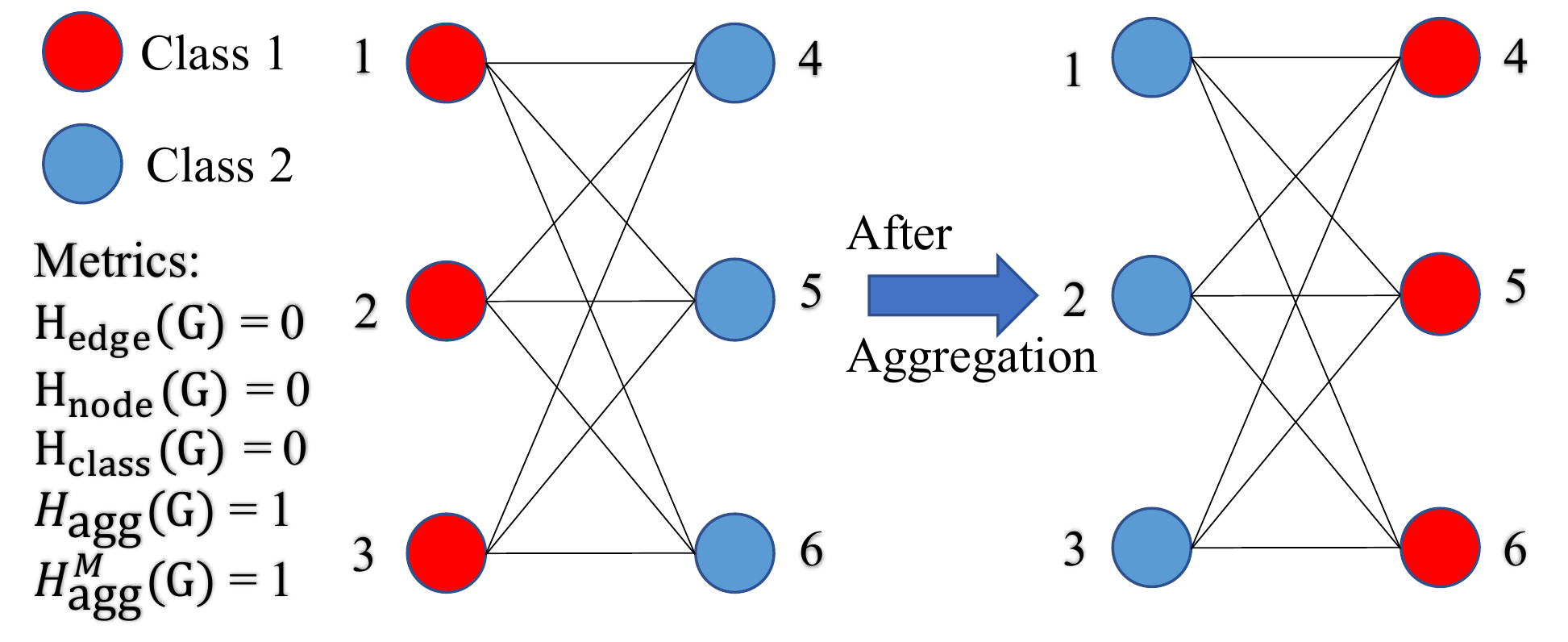}
  \end{center}
  \caption{Example of harmless heterophily}
  \label{fig:example_harmless_heterophily}
\end{wrapfigure}

\subsection{Analysis of Heterophily and Aggregation Homophily Metric}
\label{sec:heterophily_analysis}
Heterophily is believed to be harmful for message-passing based GNNs \cite{zhu2020beyond,pei2020geom,chien2021adaptive} because intuitively features of nodes in different classes will be falsely mixed and this will lead nodes indistinguishable \cite{zhu2020beyond}.
Nevertheless, it is not always the case, \eg{} the bipartite graph shown in Figure \ref{fig:example_harmless_heterophily} is highly heterophilous according to the homophily metrics in \eqref{eq:definition_homophily_metrics}, but after mean aggregation, the nodes in classes 1 and 2 only exchange colors and are still distinguishable. Authors in \cite{chien2021adaptive} also point out the insufficiency of $H_\text{node}$ by examples to show that different graph typologies with the same $H_\text{node}$ can carry different label information. 

To analyze to what extent the graph structure can affect the output of a GNN, we first simplify the GCN by removing its nonlinearity as \cite{wu2019simplifying}. Let $\hat{A} \in \mathbb{R}^{N\times N}$ denote a general aggregation operator. Then, equation \eqref{eq:gcn_original} can be simplified as,
\begin{equation}
    \begin{aligned}
    Y  = \text{softmax} (\hat{A}  X W ) = \text{softmax} (Y')
    \end{aligned}
\end{equation}
After each gradient decent step $\Delta W = \gamma \frac{d \mathcal{L}}{d W}$, where $\gamma$ is the learning rate, the update of $Y'$ will be,
\begin{equation}
    \begin{aligned}
    \label{eq:gradient_descent_update}
    \Delta Y' = \hat{A} X \Delta W = \gamma \hat{A}X \frac{d \mathcal{L}}{d W} \propto \hat{A}X \frac{d \mathcal{L}}{d W} = \hat{A}X X^T\hat{A}^T (Z-Y) = S(\hat{A},X) (Z-Y)
    \end{aligned}
\end{equation}
where $S(\hat{A},X) \equiv \hat{A}X (\hat{A}X)^T$ is a post-aggregation node similarity matrix, $Z-Y$ is the prediction error matrix. The update direction of node $i$ is essentially a weighted sum of the prediction error, \ie{} $\Delta (Y')_{i,:} = \sum_{j\in \mathcal{V}} \left[S(\hat{A},X)\right]_{i,j} (Z-Y)_{j,:}$. 

To study the effect of heterophily, we first define the {\em aggregation similarity score} as follows.

\begin{definition} Aggregation similarity score 
\begin{equation}
\label{eq:aggregation_similarity}
    S_\text{agg}\left(S(\hat{A},X)\right) = \frac{\left| \left\{v   \,\big| \,
    \mathrm{Mean}_u\big( \{S(\hat{A},X)_{v,u} | Z_{u,:}=Z_{v,:} \}\big) 
    \geq \mathrm{Mean}_u\big(\{S(\hat{A},X)_{v,u} | Z_{u,:} \neq Z_{v,:} \} \big) \right\} \right|}{\left| \mathcal{V} \right|}
\end{equation}
where $\mathrm{Mean}_u\left(\{\cdot\}\right)$ takes the average over $u$ of a given multiset of values or variables.
\end{definition}
$S_\text{agg}(S(\hat{A},X))$ measures the proportion of nodes $v\in\mathcal{V}$ that will put relatively larger similarity weights on nodes in the same class than in other classes after aggregation. It is easy to see that $S_\text{agg}(S(\hat{A},X)) \in [0,1]$. But in practice, we observe that in most datasets, we will have $S_\text{agg}(S(\hat{A},X)) \geq 0.5$. Based on this observation, we rescale \eqref{eq:aggregation_similarity} to the following modified aggregation similarity for practical usage,
\begin{equation}
  S^M_\text{agg}\left(S(\hat{A},X)\right) = \left[2 S_\text{agg}\left(S(\hat{A},X)\right)-1\right]_{+}
\end{equation}
In order to measure the consistency between labels and graph structures without considering node features and make a fair comparison with the existing homophily metrics in \eqref{eq:definition_homophily_metrics}, we define the graph ($\mathcal{G}$) aggregation ($\hat{A}$) homophily and its modified version as
\begin{equation}
    \label{eq:aggregation_homophily_metrics}
    H_{\text{agg}}(\mathcal{G}) = S_\text{agg}\left(S(\hat{A},Z)\right), \; H_{\text{agg}}^M(\mathcal{G}) = S_\text{agg}^M\left(S(\hat{A},Z)\right)
\end{equation}
In practice, we will only check $H_{\text{agg}}(\mathcal{G})$ when $H_{\text{agg}}^M(\mathcal{G})=0$. As Figure \ref{fig:example_harmless_heterophily} shows, when $\hat{A} = \hat{A}_\text{rw}$, $H_{\text{agg}}(\mathcal{G}) = H_{\text{agg}}^M(\mathcal{G}) = 1$. Thus, this new metric reflects the fact that nodes in classes 1 and 2 are still highly distinguishable after aggregation,
while other metrics mentioned before fail to capture the information and misleadingly give value 0. This shows the advantage of $H_{\text{agg}}(\mathcal{G})$ and $H_{\text{agg}}^M(\mathcal{G})$ by additionally considering information from aggregation operator $\hat{A}$ and the similarity matrix.

\begin{figure}[h]
     {
     \subfloat[$H_\text{edge}(\mathcal{G})$]{
     \captionsetup{justification = centering}
     \includegraphics[height=0.15\textheight]{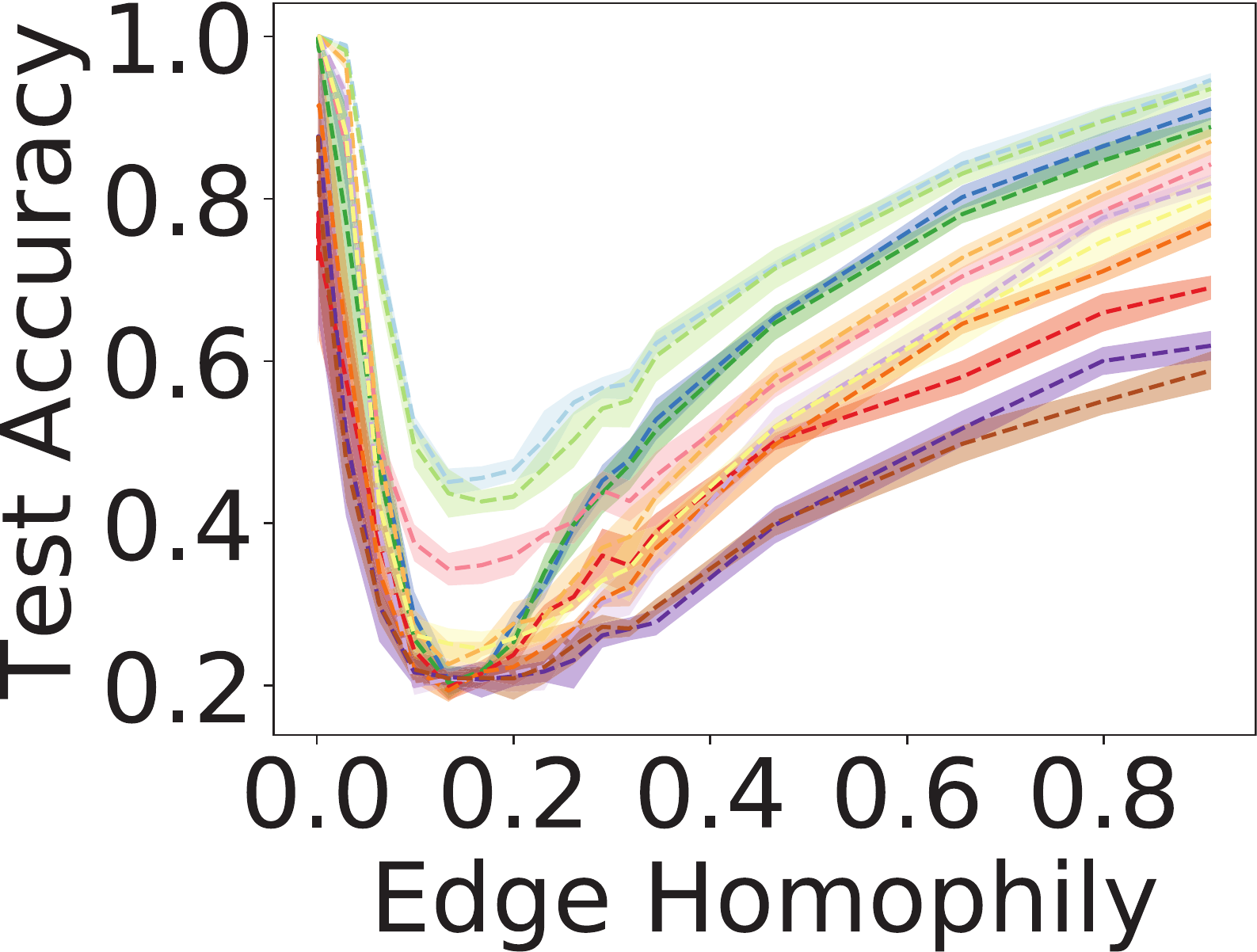}
     } 
     \subfloat[$H_\text{node}(\mathcal{G})$]{
     \captionsetup{justification = centering}
     \includegraphics[height=0.15\textheight]{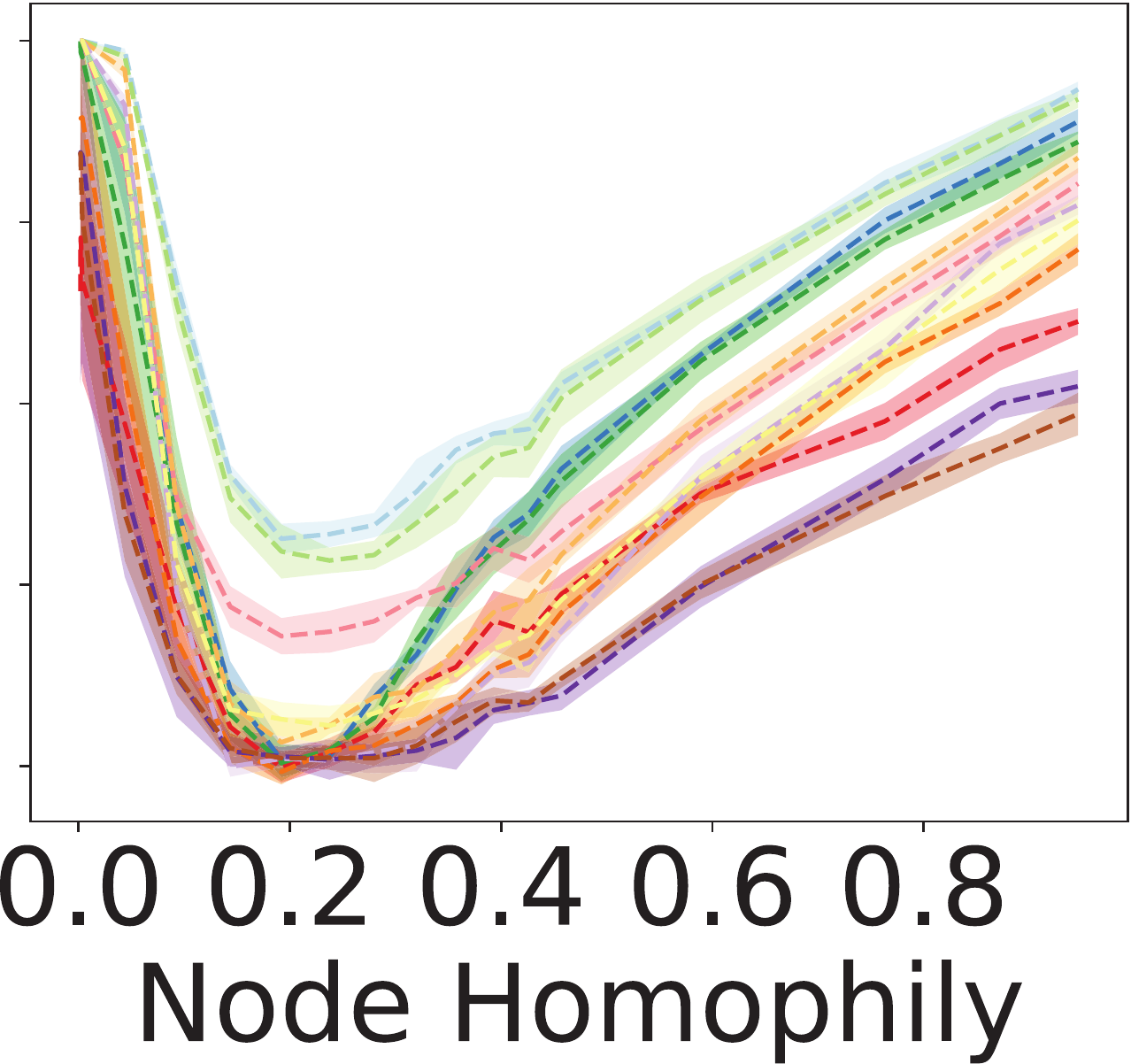}
     } 
     \subfloat[$H_\text{class}(\mathcal{G})$]{
     \captionsetup{justification=centering}
     \includegraphics[height=0.15\textheight]{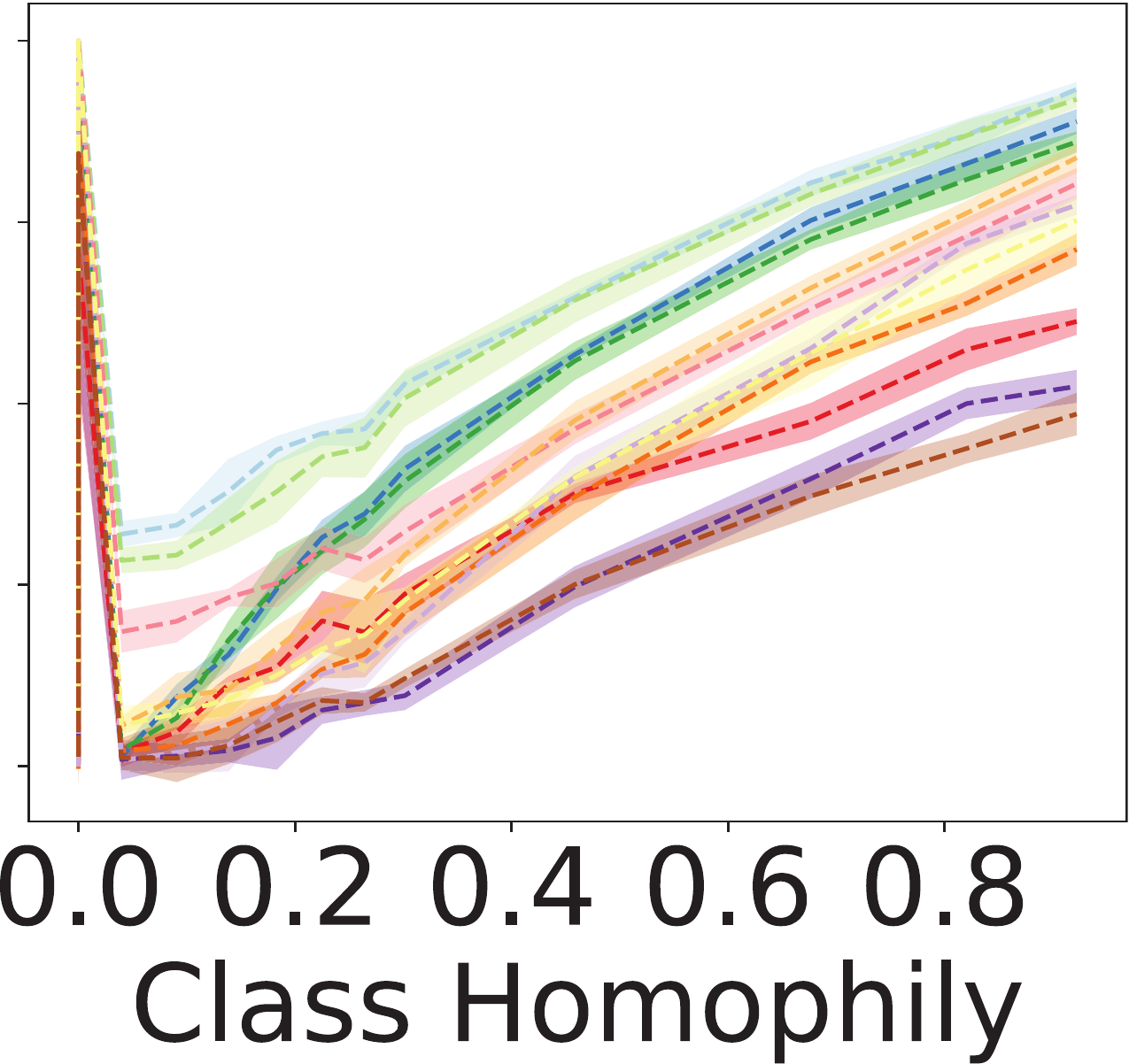}
     } 
     \subfloat[$H_{\text{agg}}^M(\mathcal{G})$]{
     \captionsetup{labelsep=newline,format=plain,indention=100pt}
     \includegraphics[height=0.15\textheight]{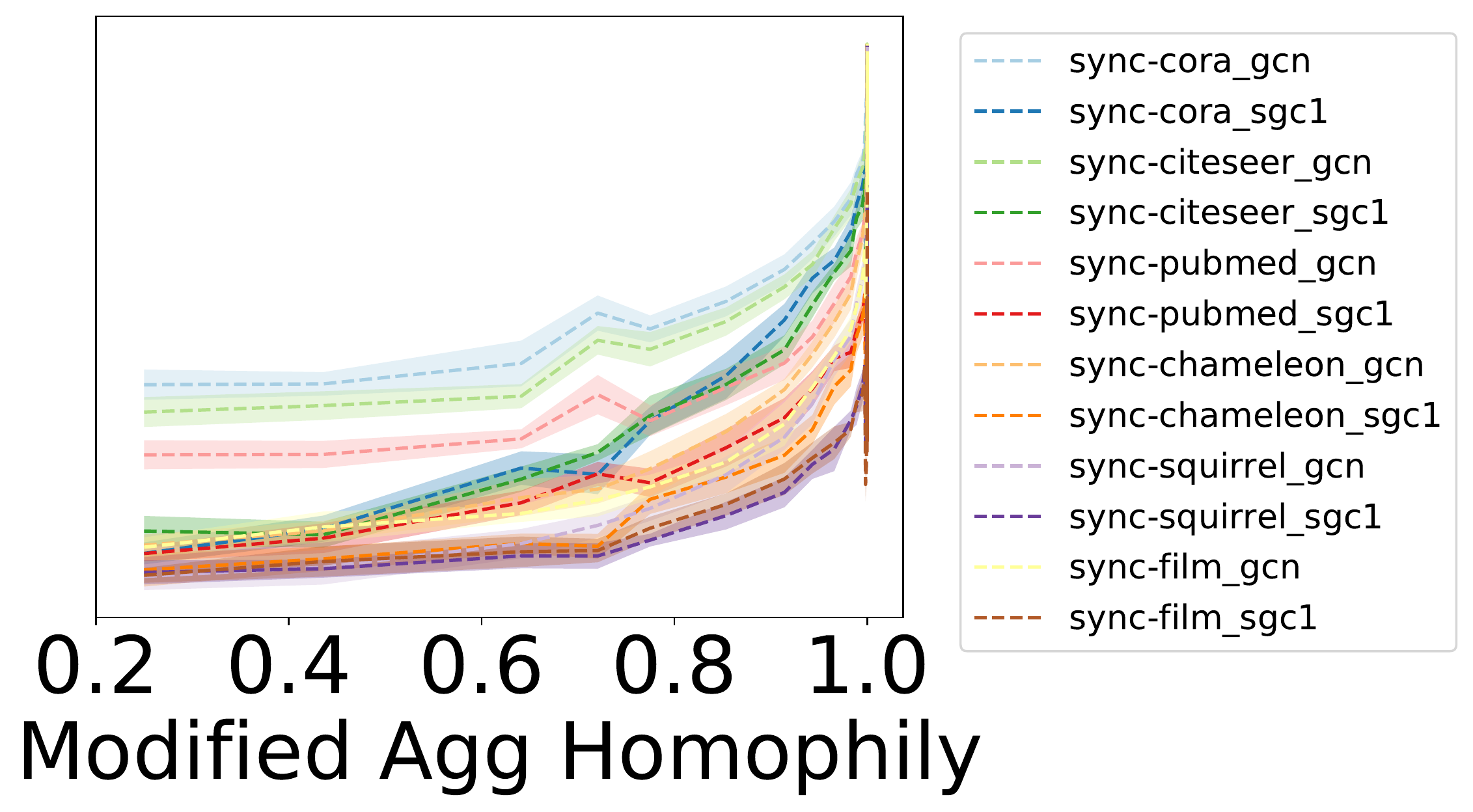}
     } 
     }
     \caption{Comparison of baseline performance under different homophily metrics.}
     \label{fig:comparison_homophily_metrics}
\end{figure}

\paragraph{Comparison of Homophily Metrics on Synthetic Graphs}
To comprehensively compare $H_{\text{agg}}^M(\mathcal{G})$ with the metrics in \eqref{eq:definition_homophily_metrics} in terms of how they reveal the influence of graph structure on the GNN performance, we generate synthetic graphs ($d$-regular graphs with edge homophily varied from $0.005$ to $0.95$) and evaluate SGC with 1-hop aggregation (SGC-1) \cite{wu2019simplifying} and GCN \cite{kipf2016classification} on them.

The performance of SGC-1 and GCN are expected to be monotonically increasing with a proper and informative homophily metric. However, Figure \ref{fig:comparison_homophily_metrics}(a)(b)(c) show that the performance curves under $H_{\text{edge}}(\mathcal{G}), H_{\text{node}}(\mathcal{G})$ and $H_{\text{class}}(\mathcal{G})$ are $U$-shaped \footnote{A similar J-shaped curve is found in \cite{zhu2020beyond}, though using different data generation processes. It does not mention the insufficiency of edge homophily.}, while Figure \ref{fig:comparison_homophily_metrics}(d) reveals a nearly monotonic curve only with a little numerical perturbation around 1. This indicates that $H_{\text{agg}}^M(\mathcal{G})$ can describe how the graph structure affects the performance of SGC-1 and GCN more appropriately and adequately than the existing metrics.

\subsection{How Diversification Operation Helps with Harmful Heterophily}
\label{sec:diversification_distinguishability}
\begin{figure}[htbp]
\centering
{
\captionsetup{justification = centering}
\includegraphics[width=1\textwidth]{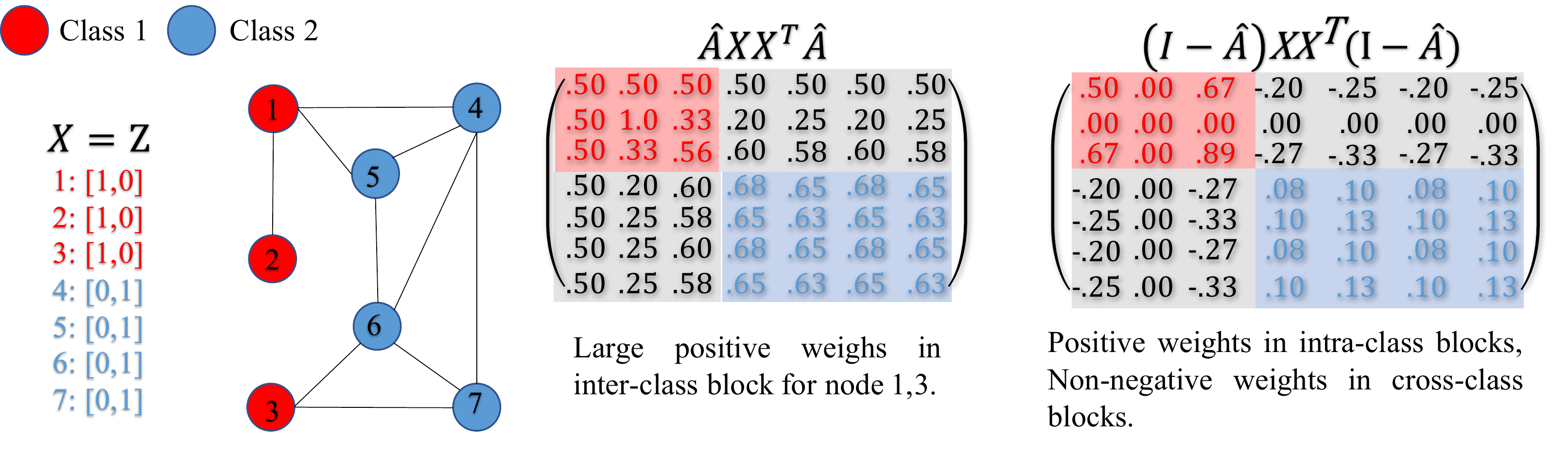}}
{%
  \caption{Example of how HP filter addresses harmful heterophily}%
  \label{fig:successful_example_hp_filter}
}
\end{figure}

We first consider the example shown in Figure \ref{fig:successful_example_hp_filter}. From $S(\hat{A},X)$, nodes 1,3 assign relatively large positive weights to nodes in class 2 after aggregation, which will 
make node 1,3 hard to be distinguished from nodes in class 2. Despite the fact, we can still 
distinguish between nodes 1,3 and 4,5,6,7 by considering 
their neighborhood difference: nodes 1,3 are different from most of their neighbors while nodes 4,5,6,7 are similar to most of their neighbors. This indicates, in some cases, although some nodes become similar after aggregation, they are still distinguishable via their surrounding dissimilarities. This leads us to use \textit{diversification operation}, \ie{} high-pass (HP) filter $I-\hat{A}$ \cite{ekambaram2014graph} (will be introduced in the next subsection) to extract the information of neighborhood differences and address harmful heterophily. As $S(I-\hat{A},X)$ in Figure \ref{fig:successful_example_hp_filter} shows, nodes 1,3 will assign negative weights to nodes 4,5,6,7 after diversification operation, 
\ie{} nodes 1,3 treat nodes 4,5,6,7 as negative samples and will move away from them during backpropagation. Base on this example, we first propose diversification distinguishability as follows to measures the proportion of nodes that diversification operation is potentially helpful for, 
\begin{definition} Diversification Distinguishability (DD) based on $S(I-\hat{A},X)$.

Given $S(I-\hat{A},X)$, a node $v$ is diversification distinguishable if the following two conditions are satisfied at the same time,
\vspace*{-1mm}
\begin{equation}
\label{eq:diversification_distinguishability}
\begin{split}
    \textbf{1.}\ \mathrm{Mean}_u \left(\{S(I-\hat{A},X)_{v,u}|u \in \mathcal{V} \land Z_{u,:}=Z_{v,:}\} \right) > 0; \\
    \textbf{2.}\ \mathrm{Mean}_u \left(\{S(I-\hat{A},X)_{v,u}|u \in \mathcal{V} \land Z_{u,:} \neq Z_{v,:}\} \right) \leq 0
    \end{split}
\end{equation}
Then, graph diversification distinguishability value is defined as
\begin{equation}
    \mathrm{DD}_{\hat{A},X}(\mathcal{G}) = \frac{1}{\left|\mathcal{V}\right|}\, \Big| \, \{v|v \mbox{ is diversification distinguishable}\}\Big|
\end{equation}
\end{definition}
\vspace*{-1mm}
We can see that $\mathrm{DD}_{\hat{A},X}(\mathcal{G}) \in [0,1]$ . The effectiveness of diversification operation can be proved for binary classification problems under certain conditions based on definition $2$, leading us to:
\begin{theorem} 3
Suppose $X=Z, \hat{A}=\hat{A}_{\text{rw}}$.
Then, for a binary classification problem, \ie{} $C=2$, all nodes are diversification distinguishable, \ie{} $\mathrm{DD}_{\hat{A},Z}(\mathcal{G})=1$.
\end{theorem}
Theorem 3 theoretically demonstrates the importance of diversification operation to extract high-frequency information of graph signal \cite{ekambaram2014graph}. Combined with aggregation operation, which is a low-pass filter \cite{ekambaram2014graph,maehara2019revisiting}, we can get a filterbank which uses both aggregation and diversification operations to distinctively extract the low- and high-frequency information from graph signals. We will introduce filterbank in the next subsection.

\subsection{Filterbank and Adaptive Channel Mixing(ACM) GNN  Framework}
\label{sec:acm_gnn_architecture}
\paragraph{Filterbank} For the graph signal $\bm{x}$ defined on $\mathcal{G}$, a 2-channel linear (analysis) filterbank \cite{ekambaram2014graph} \footnote{In graph signal processing, an additional synthesis filter \cite{ekambaram2014graph} is required to form the 2-channel filterbank. But synthesis filter is not needed in our framework, so we do not introduce it in our paper.} includes a pair of low-pass(LP) and high-pass(HP) filters $H_\text{LP}, H_\text{HP}$, where $H_\text{LP}$ and $H_\text{HP}$ retain the low-frequency and high-frequency content of $\bm{x}$, respectively. Filterbanks with $H_\text{LP} + H_\text{HP} = I$ will not lose any information of the input signal, \ie{} perfect reconstruction property \cite{ekambaram2014graph}.

However, most existing GNNs are under uni-channel filtering architecture \cite{kipf2016classification,velivckovic2017attention,hamilton2017inductive} with either $H_\text{LP}$ or $H_\text{HP}$ channel that only partially preserves the input information. Generally, the Laplacian matrices ($L_\text{sym}$, $L_\text{rw}$, $\hat{L}_\text{sym}$, $\hat{L}_\text{rw}$) can be regarded as HP filters \cite{ekambaram2014graph} and affinity matrices ($A_\text{sym}$, $A_\text{rw}$, $\hat{A}_\text{sym}$, $\hat{A}_\text{rw}$) can be treated as LP filters \cite{maehara2019revisiting, hamilton2020graph}. Moreover, we consider MLPs as owing a special identity filterbank with matrix $I$ that satisfies $H_\text{LP} + H_\text{HP} = I+0 = I$.
\paragraph{Filterbank in Spatial Form}
Filterbank methods can also be extended to spatial GNNs. Formally, on the node level, left multiplying $H_\text{LP}$ and $H_\text{HP}$ on $\bm{x}$ performs as aggregation and diversification operations, respectively. For example, suppose $H_\text{LP} = \hat{A}$ and $H_\text{HP}=I-\hat{A}$, then for node $i$ we have
\begin{equation}
\label{eq:spatial_form_filterbank}
    (H_\text{LP} \bm{x})_i =  \sum_{j \in \{\mathcal{N}_i \cup i\} } \hat{A}_{i,j} \bm{x_j}, \ (H_\text{HP} \bm{x})_i = \bm{x_i} - \sum_{j \in \{\mathcal{N}_i \cup i\} } \hat{A}_{i,j} \bm{x_j}
\end{equation} 
where $\hat{A}_{i,j}$ is the connection weight between two nodes. To leverage HP and identity channels in GNNs, we propose the Adaptive Channel Mixing (ACM) framework which can be applied to lots of baseline GNN. We use GCN as an example and introduce ACM framework in matrix form. We use $H_\text{LP}$ and $H_\text{HP}$ to represent general LP and HP filters. The ACM framework includes $3$ steps as follows,
\vspace*{-1.5mm}
\begin{equation}
\begin{aligned}
\label{eq:acm_gnn_spectral}
& \textbf{{Step 1. Feature Extraction for Each Channel:}} \\
& \text{Option 1: } {H}^{l}_L = \text{ReLU}\left(H_\text{LP} {H^{l-1}} W^{l-1}_L\right), \ {{H}^{l}_H} =  \text{ReLU}\left(H_\text{HP} {H^{l-1}} W^{l-1}_H\right), {H}^{l}_I  = \ \text{ReLU}\left(I {H^{l-1}} W^{l-1}_I\right); \\
& \text{Option 2: } {H}^{l}_L = H_\text{LP} \text{ReLU}\left({H^{l-1}} W^{l-1}_L\right), \ {{H}^{l}_H} = H_\text{HP} \text{ReLU}\left({H^{l-1}} W^{l-1}_H\right), {H}^{l}_I  = I\ \text{ReLU}\left({H^{l-1}} W^{l-1}_I\right); \\
& W_L^{l-1},\ W_H^{l-1}, \ W_I^{l-1} \in \mathbb{R}^{F_{l-1} \times F_l}; \\
& \textbf{Step 2. Feature-based Weight Learning} \\
&\tilde{\alpha}_L^l = \sigma\left({H}^{l}_L \tilde{W}^{l}_L\right),\ \tilde{\alpha}_H^l = \sigma \left({H}^{l}_H \tilde{W}^{l}_H\right),\ \tilde{\alpha}_I^l = \sigma \left({H}^{l}_I \tilde{W}^{l}_I\right),\ \tilde{W}_L^{l-1},\ \tilde{W}_H^{l-1},\ \tilde{W}_I^{l-1} \in \mathbb{R}^{F_l \times 1}\\ 
& \left[{\alpha}_L^l, {\alpha}_H^l, {\alpha}_I^l \right] = \text{Softmax}\left(\left[\tilde{\alpha}_L^l,\tilde{\alpha}_H^l,\tilde{\alpha}_I^l\right]W_\text{Mix}^l/T, \right),\ W_\text{Mix}^l \in \mathbb{R}^{3\times 3}, T \in \mathbb{R} \text{ is the temperature} ; \\
&\textbf{Step 3. Node-wise Channel Mixing:}\\
& {H^{l}}  = \left( \text{diag}(\alpha_L^l){H}^{l}_L + \text{diag}(\alpha_H^l){H}^{l}_H + \text{diag}(\alpha_I^l){H}^{l}_I \right).
\end{aligned}
\end{equation}
The framework with option 1 in step 1 is ACM framework and with option 2 is ACMII framework. ACM(II)-GCN first implement distinct feature extractions for $3$ channels, respectively. After processed by a set of filterbanks, $3$ filtered components $H_L^l,H_H^l, H_I^l$ are obtained. Different nodes may have different needs for the information in the 3 channels, \eg{} in Figure \ref{fig:successful_example_hp_filter}, nodes 1,3 demand high-frequency information while node 2 only needs low-frequency information. To adaptively exploit information from different channels, ACM(II)-GCN learns row-wise (node-wise) feature-conditioned weights to combine the $3$ channels. ACM(II) can be easily plugged into spatial GNNs by replacing $H_\text{LP}$ and $H_\text{HP}$ by aggregation and diversification operations as shown in \eqref{eq:spatial_form_filterbank}.
\vspace*{-2.5mm}
\paragraph{Complexity} Number of learnable parameters in layer $l$ of ACM(II)-GCN is $3F_{l-1}(F_l +1)+9$, while it is $F_{l-1}F_l$ in GCN. The computation of step 1-3 takes $NF_l(8+6F_{l-1}) + 2F_l(\text{nnz}(H_\text{LP}) + \text{nnz}(H_\text{HP}))+ 18N$ flops, while GCN layer takes $2NF_{l-1}F_l + 2F_l(\text{nnz}(H_\text{LP}))$ flops, where $\text{nnz}(\cdot)$ is the number of non-zero elements.

\begin{figure}[h]
     {
     \subfloat[$H_\text{agg}^M(\mathcal{G})=0.8032$, $\uparrow$ 4.1 \%]{
     \captionsetup{justification = centering}
     \includegraphics[height=0.15\textheight]{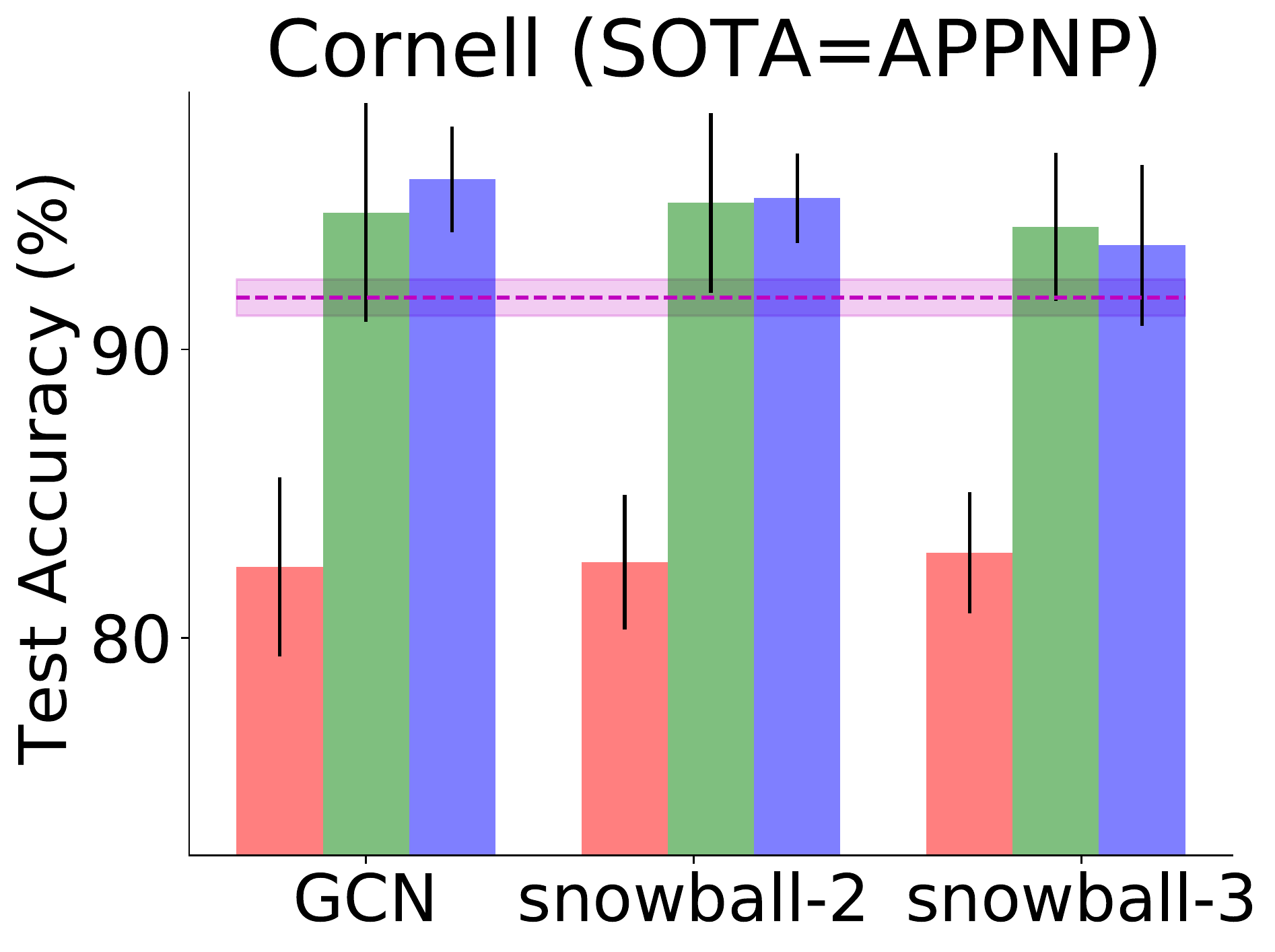}
     } 
     \subfloat[$H_\text{agg}^M(\mathcal{G})=0.7768$, $\uparrow$3.13 \%]{
     \captionsetup{justification = centering}
     \includegraphics[height=0.15\textheight]{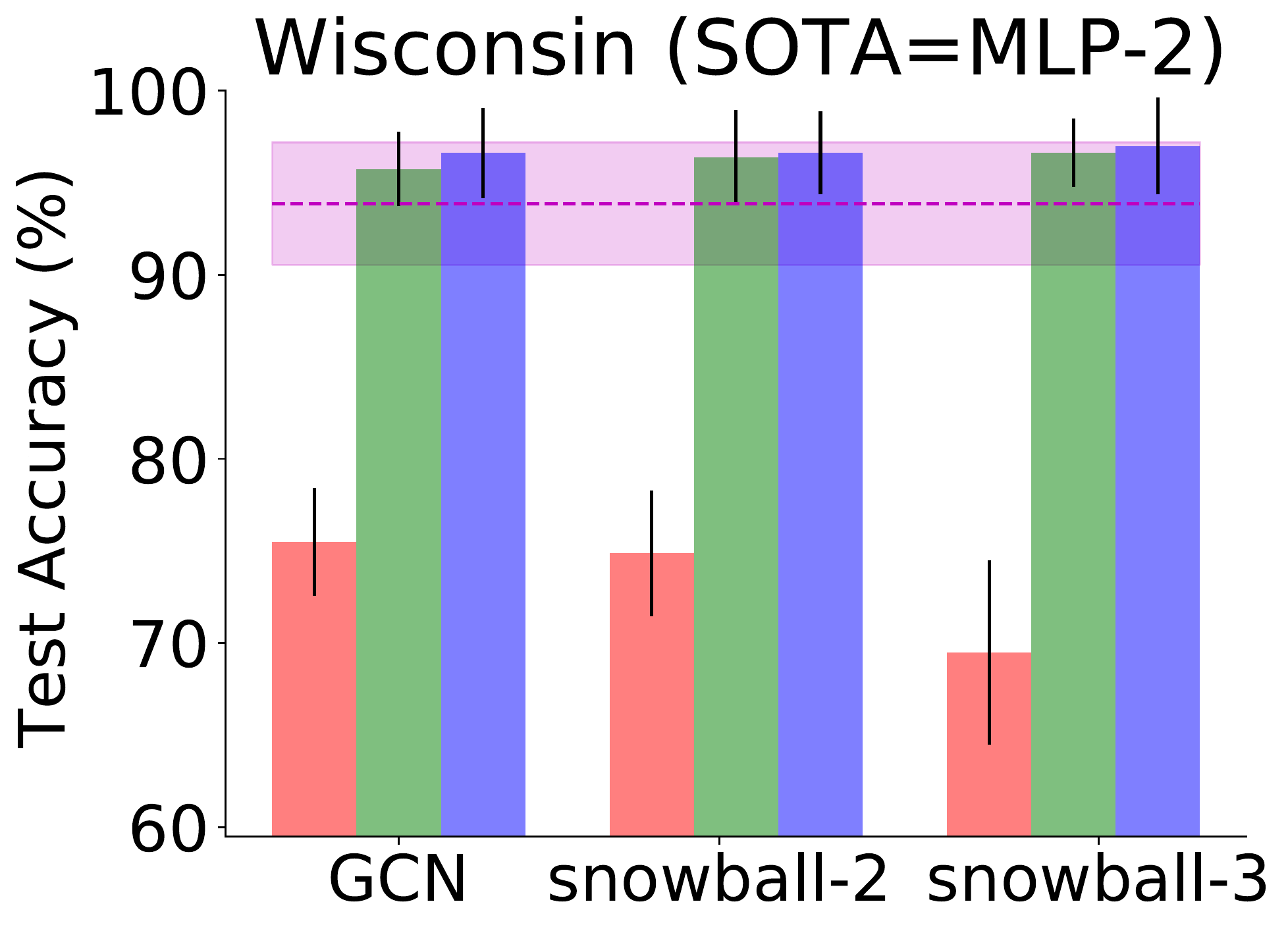}
     } 
     \subfloat[$H_\text{agg}^M(\mathcal{G})=0.694$, $\uparrow$2.82 \%]{
     \captionsetup{justification=centering}
     \includegraphics[height=0.15\textheight]{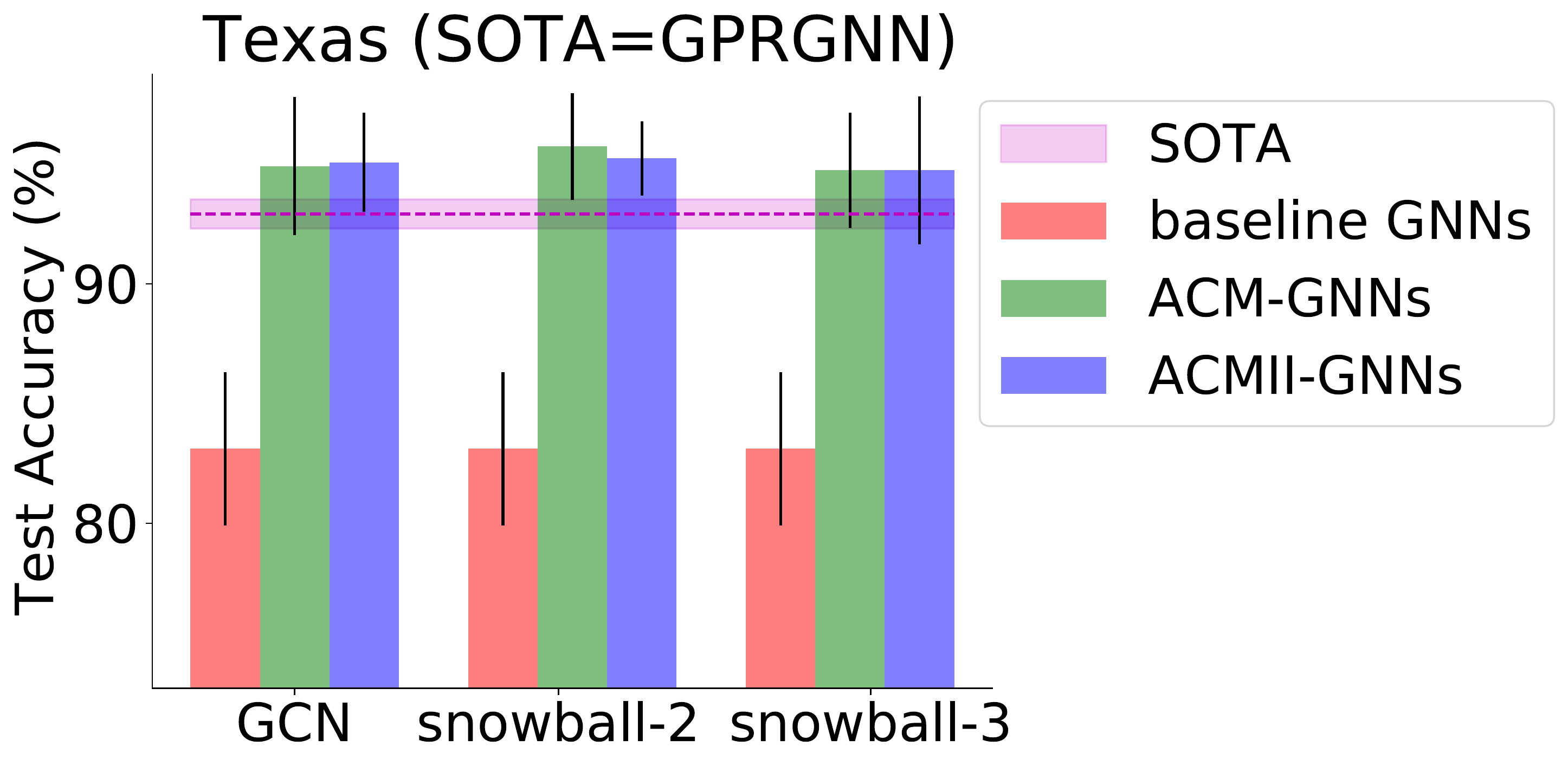}
     }\\
    \subfloat[$H_\text{agg}^M(\mathcal{G})=0.61$, $\uparrow$ 0.9 \%]{
     \captionsetup{justification = centering}
     \includegraphics[height=0.15\textheight]{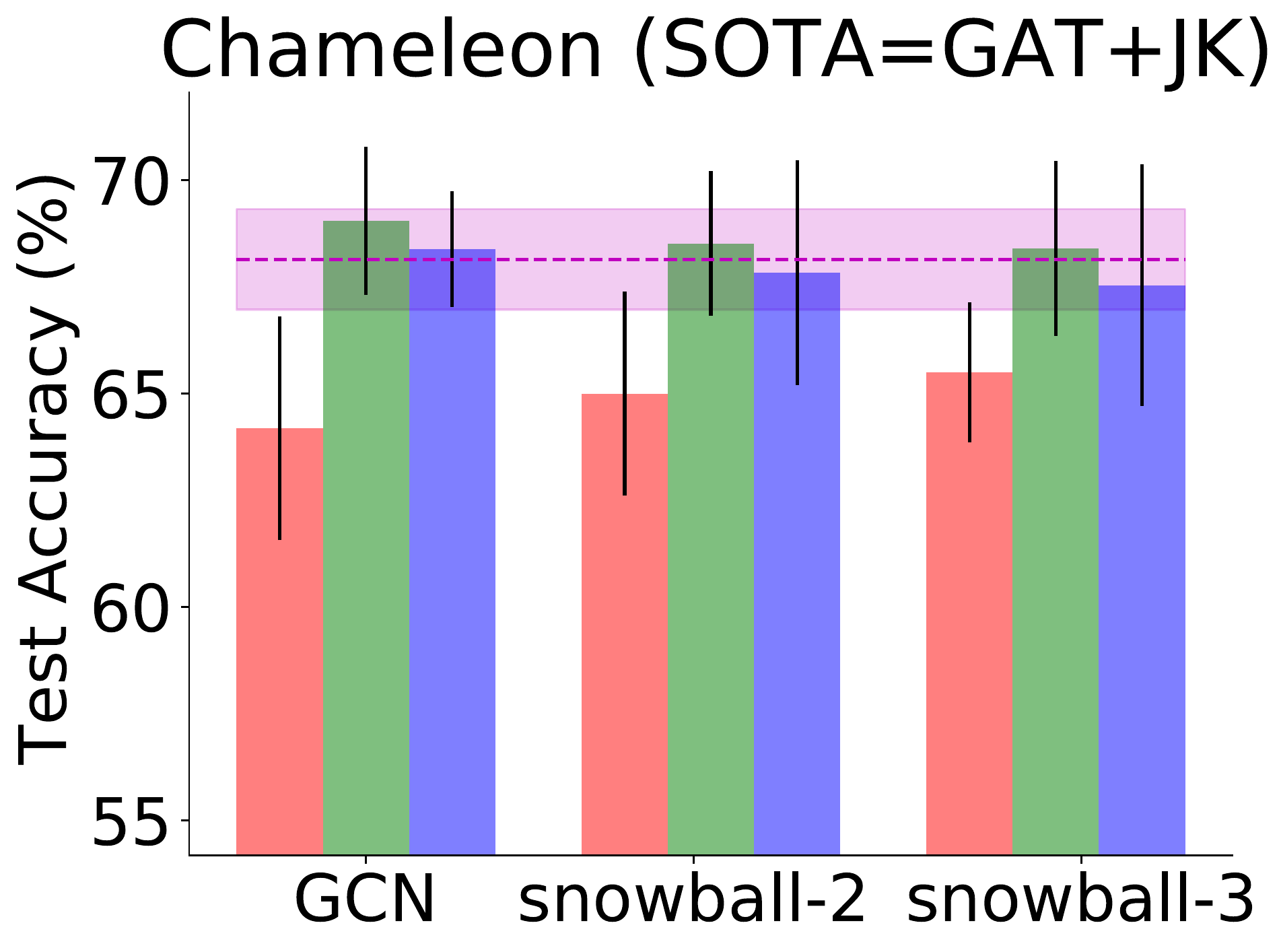}
     } 
     \subfloat[$H_\text{agg}^M(\mathcal{G})=0.6822$, $\uparrow$2.54\%]{
     \captionsetup{justification = centering}
     \includegraphics[height=0.15\textheight]{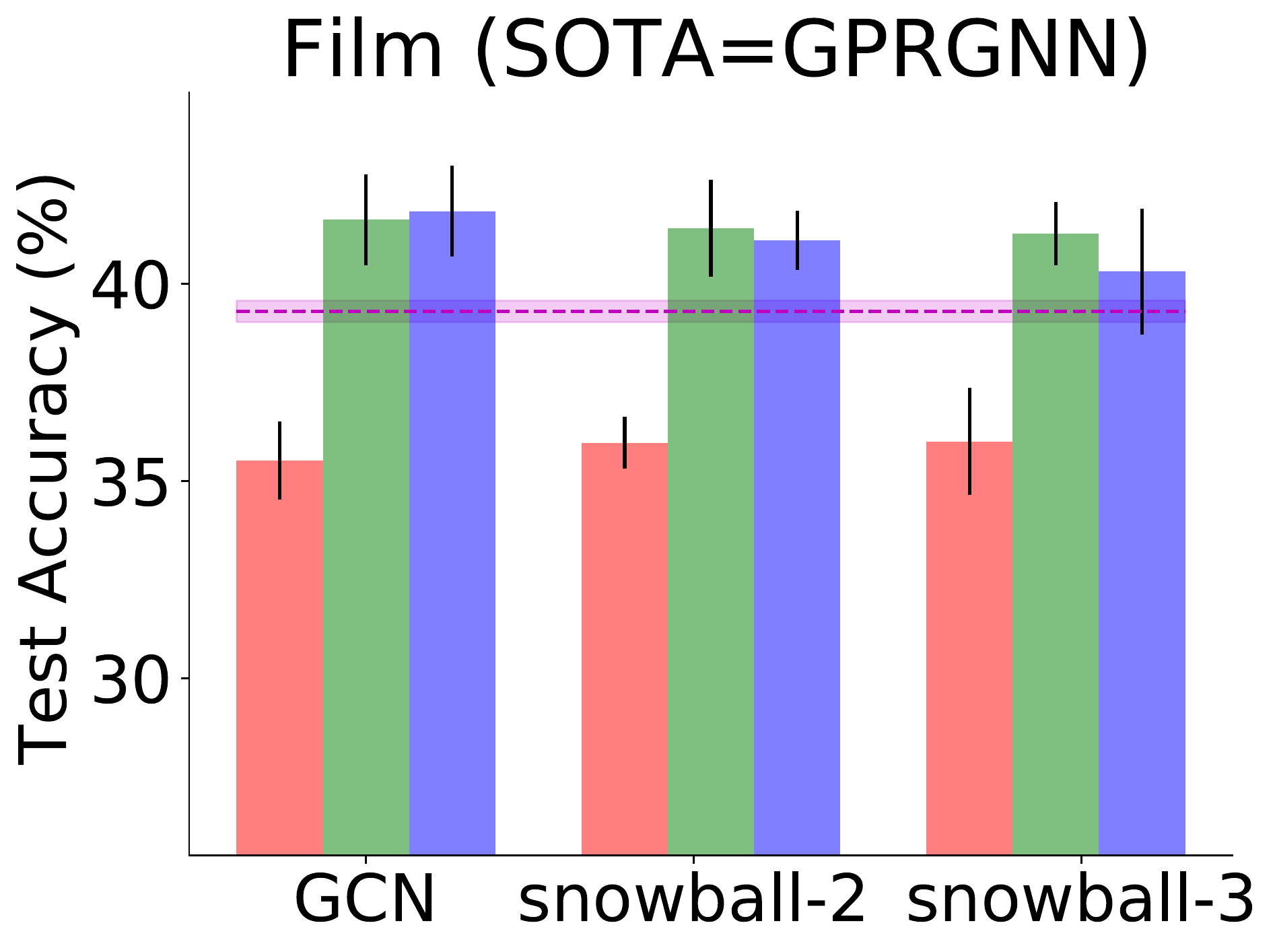}
     } 
     \subfloat[$H_\text{agg}^M(\mathcal{G})=0.3566$, $\uparrow$4.62 \%]{
     \captionsetup{justification=centering}
     \includegraphics[height=0.15\textheight]{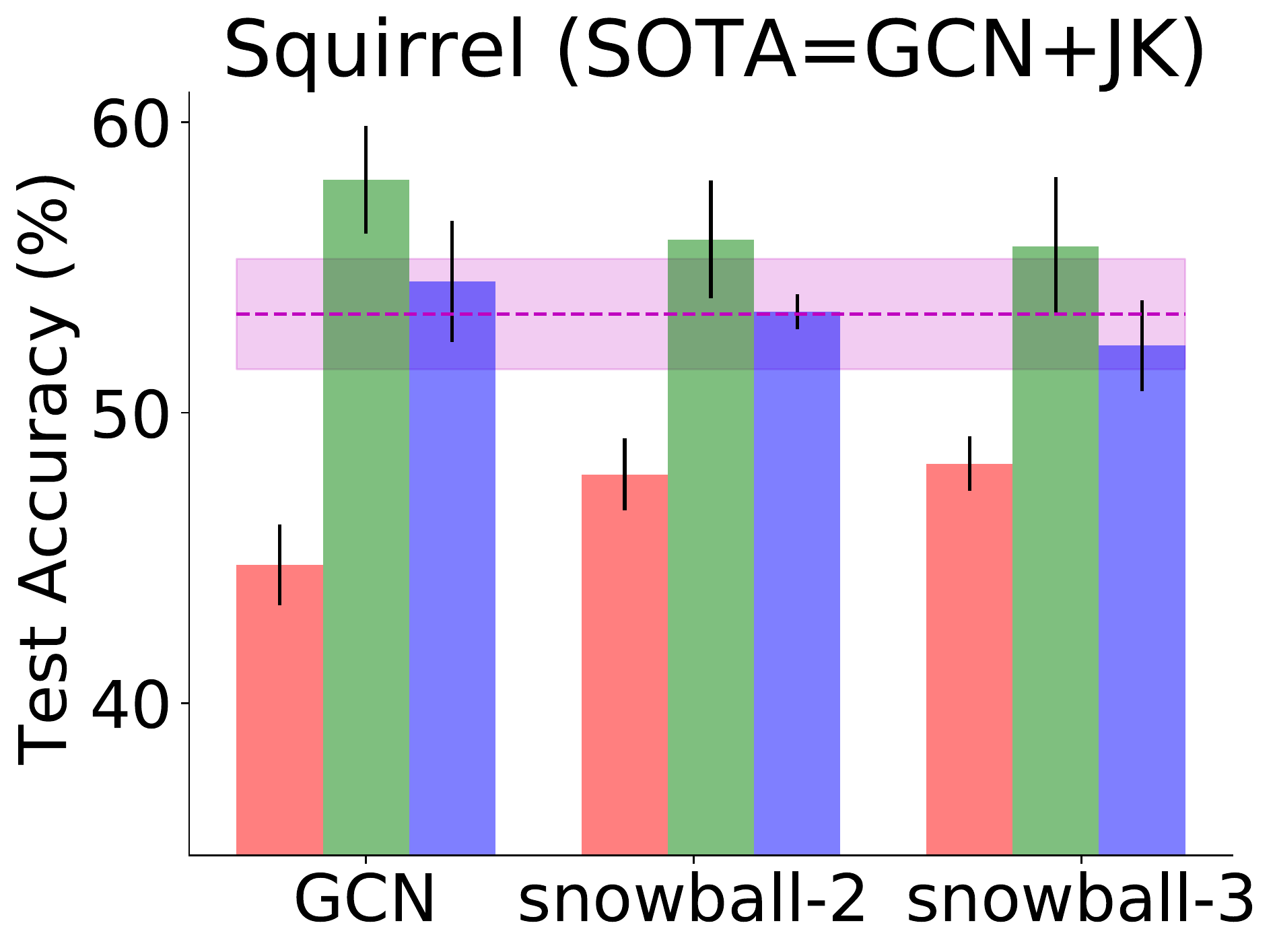}
     }
     }
     \caption{Comparison of SOTA models (magenta), selected baseline GNNs (red) and their ACM (green) and ACMII (blue) augmented models on $6$ selected datasets. The black line and the error bar indicate the standard deviation. The symbol “$\uparrow$” means the amount of improvement of the best ACM-baseline and ACM-baseline over the SOTA models. }
     \label{fig:selected_comparison_with_sota}
\end{figure}
\paragraph{Performance Comparison}
We implement SGC \cite{wu2019simplifying} with 1 hop and 2 hops (SGC-1, SGC-2), GCNII \cite{chen2020simple}, GCNII$^*$ \cite{chen2020simple}, GCN \cite{kipf2016classification} and snowball networks with 2 and 3 layers (snowball-2, snowball-3) and apply them in ACM or ACMII framework: we use $\hat{A}_\text{rw}$ as LP filter and the corresponding HP filter is $I-\hat{A}_\text{rw}$. We compare them with several baseline and SOTA GNN models: MLP with 2 layers (MLP-2), GAT \cite{velivckovic2017attention},  APPNP \cite{klicpera2018predict}, GPRGNN \cite{chien2021adaptive}, H$_2$GCN \cite{zhu2020beyond}, MixHop \cite{abu2019mixhop}, GCN+JK \cite{kipf2016classification, pmlr-v80-xu18c, lim2021new}, GAT+JK \cite{velivckovic2017attention, pmlr-v80-xu18c, lim2021new}, FAGCN \cite{bo2021beyond} GraphSAGE \cite{hamilton2017inductive} and Geom-GCN \cite{pei2020geom}. Besides the 9 benchmark datasets  \textit{Cornell}, \textit{Wisconsin}, \textit{Texas}, \textit{Film}, \textit{Chameleon}, \textit{Squirrel}, \textit{Cora}, \textit{Citeseer} and \textit{Pubmed}  used in \cite{pei2020geom}, we further test the above models on a new benchmark dataset, \textit{Deezer-Europe}, that is proposed in \cite{lim2021new}.
On each dataset used in \cite{pei2020geom}, we test the models 10 times following the same early stopping strategy, the same random data splitting method and Adam \cite{kingma2014adam} optimizer as used in GPRGNN \cite{chien2021adaptive}.  
For \textit{Deezer-Europe}, we test the above models 5 times with the same early stopping strategy, the same fixed splits and AdamW \cite{loshchilov2017decoupled} used in \cite{lim2021new}. 

To better visualize the performance boost and the comparison with SOTA models, in Figure \ref{fig:selected_comparison_with_sota}, we plot the bar charts of the test accuracy of SOTA models, $3$ selected baselines (GCN, snowball-2, snowball-3) and their ACM and ACMII augmented models on $6$ most commonly used benchmark heterophily datasets (See \cite{luan2021heterophily2} for the full results and comparison). We can see that after being applied in ACM or ACMII framework, the performance of the $3$ baseline models are significantly boosted on all tasks and can achieve SOTA performance. Especially on \textit{Cornell, Texas, Film} and \textit{Squirrel}, the augmented models significantly outperform the current SOTA models. Overall, It suggests that ACM or ACMII framework can help GNNs to generalize better on node classification tasks on heterophilous graphs.

\subsection{Prior Work}
\label{sec:heterophily_prior_work}
We discuss relevant work of GNNs on addressing heterophily challenge in this part. Authors in \cite{abu2019mixhop} acknowledge the difficulty of learning on graphs with weak homophily and propose MixHop to extract features from multi-hop neighborhood to get more information. Geom-GCN \cite{pei2020geom} precomputes unsupervised node embeddings and uses graph structure defined by geometric relationships in the embedding space to define the bi-level aggregation process. Authors in \cite{hou2019measuring} propose measurements based on feature smoothness and label smoothness that are potentially helpful to guide GNNs on dealing with heterophilous graphs. H$_2$GCN \cite{zhu2020beyond} combines 3 key designs to address heterophily: (1) ego- and neighbor-embedding separation; (2) higher-order neighborhoods; (3) combination of intermediate representations. CPGNN \cite{zhu2020graph} models label correlations by the compatibility matrix, which is beneficial for heterophily settings, and propagates a prior belief estimation into GNNs by the compatibility matrix. FBGNN \cite{luan2022complete} first proposes to use filterbank to address heterophily problem, but it does not fully explain the insights behind HP filters and does not contain identity channel and node-wise channel mixing mechanism. FAGCN \cite{bo2021beyond} learns edge-level aggregation weights as GAT \cite{velivckovic2017attention} but allows the weights to be negative which enables the network to capture the high-frequency components in graph signals. GPRGNN \cite{chien2021adaptive} uses learnable weights that can be both positive and negative for feature propagation, it allows GRPGNN to adapt heterophily structure of graph and is able to handle both high- and low-frequency parts of the graph signals.

\subsection{Future Work}
\label{sec:adj_learner_deeper_thinking}
\paragraph{Limitation of diversification operation} Diversification operation does not work well in all harmful heterophily cases. For example, consider an imbalanced dataset where several small clusters with distinctive labels are densely connected to a large cluster. In this case, the surrounding differences of nodes in small clusters are similar, \ie{} the neighborhood differences are mainly from their connection to the same large cluster, and this possibly makes diversification operation fail to discriminate them. Thus, it is obvious that ACM framework is not able to handle all heterophily cases.

\begin{figure}[htbp]
\centering
{
\captionsetup{justification = centering}
\includegraphics[width=1\textwidth]{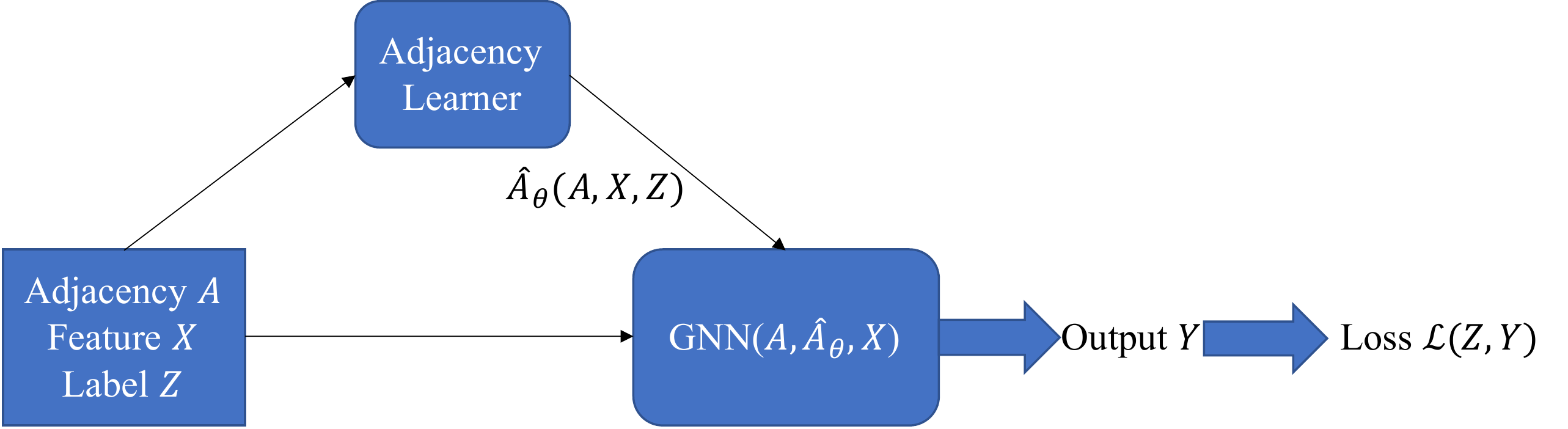}}
\caption{GNN with adjacency learner}
\label{fig:gnn_with_adjacency_learner}
\end{figure}

From Figure \ref{fig:comparison_homophily_metrics}, we can see that GNNs consistently perform well in the high homophily area. This reveals the fact that all homophily cases are helpful. This reminds us that instead of using a fixed adjacency matrix, we can learn a new adjacency matrix with different homophily level. With this in mind, we design an architecture with additional adjacency learner as shown in Figure \ref{fig:gnn_with_adjacency_learner}: instead of using a fixed predefined adjacency matrix, we will learn an adjacency matrix with edges that can reveal the label similarity between nodes, \ie{} homophily. . This adjacency learner should ideally be trained end-to end. From some preliminary experimental results (not included in this report) of a GCN with a pretrained adjacency learner, this method is promising although there are some stability issues need to be fixed.
\paragraph{Exploring Different Ways for Adjacency Candidate Selection} 
Some tricks can be explored when we are selecting the adjacency candidates for the adjacency learner:
\begin{itemize}
    \item Sample or select (top-$k_1$) nodes from complementary graph, put them  together with the pre-defined neighborhood set to form adjacency candidate set, then sample or select (top-$k_2$) adjacency candidates for training. Try to train it end-to-end.
    \item Consider modeling the candidate selection process as a multi-armed bandit problem. Find an efficient way to learn to select good candidates from complementary graph. Can use pseudo count to prevent selecting the same nodes repeatedly.
\end{itemize}


\section{Graph Representation Learning for Reinforcement Learning}
\label{sec:grl_for_rl}
\subsection{Markov Decision Process (MDP)}
MDP is a framework to model the learning process that the agent learns from the interaction with the environment\cite{sutton2018reinforcement,zhao2021consciousness,zhao2019meta}. The interaction happens in discrete time steps, $t=0,1,2,3,\cdots$. At step $t$, given a state $S_t = s_t \in \mathcal{S}$, the agent picks an action $a_t \in \mathcal{A}(s_t)$ according to a policy $\pi(\cdot | s_t)$, which is a rule of choosing actions given a state. Then, at time $t+1$, the environmental dynamics $p: \mathcal{S} \times \mathcal{R} \times \mathcal{A} \times \mathcal{S} \rightarrow [0,1]$ take the agent to a new state $S_{t+1} = s_{t+1} \in \mathcal{S}$ and provide a numerical reward $R_{t+1} = r_{t+1}(s_t, a_t, s_{t+1}) \in  \mathbb{R}$. Such a sequence of interaction gives us a trajectory $\tau = \{ S_0, A_0,R_1,S_1,A_1,R_2,S_2,A_2,R_3,\cdots\}$. The objective is to find an optimal policy to maximize the expected long-term discounted cumulative reward $V_\pi(s) = E_\pi[\sum\limits_{k=0}^{\infty} \gamma^k R_{t+k+1}|S_t=s]$ for each state $s$, where $\gamma$ is the discount factor. 

For a given policy $\pi$, solving its value function $\bm{V}_\pi$ is equivalent to solving the following linear system,
\begin{equation}
    \bm{V}_\pi = \bm{r}_\pi + \gamma P_\pi \bm{V}_\pi
\end{equation}
where $\bm{V}_\pi = [V_\pi(s)]_{s\in \mathcal{S}}^T \in \mathbb{R}^{|\mathcal{S}|}, \bm{r}_\pi = [r_\pi(s)]_{s\in \mathcal{S}}^T \in \mathbb{R}^{|\mathcal{S}|}, P_\pi = [P_\pi(s'|s)]_{s',s\in \mathcal{S}} \in \mathbb{R}^{|\mathcal{S}| \times |\mathcal{S}|}$. The state transition matrix $P_\pi$ essentially defines a graph structure over states and the reward vector $\bm{r}_\pi$ is a signal defines on graph. Thus, solving value function can be considered as a (supervised or semi-supervised) node regression tasks over graph. Besides solving $\bm{V}_\pi$, the graph structure can also be used for reward propagation and representation learning in Reinforcement Learning (RL) \cite{martin2018diffusion,martin2019graph,klissarov2020reward}.
\subsection{Graph Representation Learning for MDP}
Treating MDP as a graph is an old but never outdated idea. Traditional methods use graph Laplacian for a fixed policy to estimate $\bm{V}_\pi$, \eg{} proto-value function \cite{mahadevan2005proto}.  In addition to value function estimation, \cite{klissarov2020reward} proposes to use GCN to learn potential-based reward shaping, which can accelerate the learning process of the agent.

The above methods both construct the graph from the sampled trajectory data. With modern Graph Representation Learning (GRL) methods \eg{} node embedding methods \cite{bordes2013translating}, link prediction methods \cite{perozzi2014deepwalk,tang2015line,ribeiro2017struc2vec}, we can learn to reconstruct the underlying graph (adjacency matrix) from sampled data more efficiently. And label propagation \cite{raghavan2007near}, which is a commonly used algorithm for graph semi-supervised learning, can be helpful for efficient reward propagation. In section \ref{sec:app_rl}, we will introduce the potential of using GRL for reward propagation and representation learning in reinforcement learning.

\subsection{Reinforcement Learning with Graph Representation Learning}
\label{sec:app_rl}


In this section, we will draw how to represent Markov Decision Process (MDP) with graph and introduce two possible ways of using graph representation learning to address the problems defined on MDP. 

 Each state can be treated as a node on a graph, the transition probability between each pair of nodes (an element in state transition matrix) can be represented by the edge (or weight) between them and value function is a function defined on each node of the graph. The details (for finite MDP) are introduced in matrix form as follows \cite{wang2007dual,luan2019revisit}:
 \begin{itemize}
    \item  Denote $|S||A| \times |S|$ environment transition matrix as $P$, where 
    \begin{equation} \label{environment dynamics}
    P_{\left(s a, s^{\prime}\right)} = \sum\limits_{r} p\left(s^{\prime},r | s, a\right)   
    \end{equation}
    and $ P_{\left(s a, s^{\prime}\right)} \geq 0$, $\sum_{s^{\prime}} P_{\left(s a, s^{\prime}\right)} = 1$, for all $s, a$. Note that $P$ is not a square matrix.
    \item  We rewrite the policy $\pi$ by an $|S|\times |S||A|$ matrix $\Pi$, where $\Pi_{\left(s, s^{\prime} a\right)}=\pi(a|s) \text { if } s^{\prime}=s$, otherwise 0:
    \begin{equation} \label{policy}
    \Pi = \text{diag}(\pi(\cdot | s_1)^T, \cdots, \pi(\cdot | s_{|S|})^T)
    \end{equation}
where $\pi(\cdot | s_i)^T$ is an $|A|$-dimensional row vector. From this definition, one can quickly verify that the matrix product $\Pi P $ gives the $|S| \times|S|$ state-to-state transition matrix $P_\pi$ (asymmetric) induced by the policy $\pi$ in the environment $P$, and the $|S||A| \times|S||A|$ matrix product $P \Pi$ gives the state-action-to-state-action transition matrix $P_\pi'$ (asymmetric) induced by policy $\pi$ in the environment $P$.
    
    \item We denote the $|S||A|\times 1$ reward vector as $\bm{r}$, whose entry $r_{(sa)}$ specifies the reward obtained when taking action $a$ in state $s$, i.e.
    \begin{equation} \label{reward}
      \bm{r}_{(sa)} = E[r|s,a] = \sum\limits_{s' \in \mathcal{S}} P_{sa, s'} \cdot r(s,a,s').
    \end{equation}
    
    \item The state value function and state-action value function can be represented by
    \begin{equation} \label{value function}
          \bm{V_\pi}=\sum\limits_{i=0}^{\infty} \gamma^{i}(\Pi P)^{i} \Pi \bm{r} = \Pi \bm{r}+\gamma \Pi P \bm{V_\pi}  \in \mathbb{R}^{|S| \times 1}, \; \bm {Q_\pi} = \sum\limits_{i=0}^{\infty} \gamma^{i} (P \Pi)^{i} \bm {r} = \bm{r} + \gamma P \Pi \bm{Q_\pi} \in \mathbb{R}^{|S||A| \times 1}
\end{equation}
\end{itemize}

\subsubsection{Learn Reward Propagation as Label Propagation} 
\label{diffusion and embedding}


The sampling process  from an MDP can be considered as a random walk defined on a graph, because the relation (edge) between each pair of states is essentially a transition probability \footnote{From this perspective, we should not treat the trajectory as sequential data, because we do not necessarily have an ordered relation between states on a graph, even for directed graph. Although the observation seems to have an order in it, we actually only have transition relation.}. Discovering the underlying graph of a MDP can help us to leverage the correlation between states to learn value function or do to efficient exploration in sparse reward environment. 

Usually, the graph is constructed from the trajectory data, \ie{} the pairwise state transition data. But once we update the policy, we need to reconstruct the graph. With graph embedding methods for link prediction tasks, \eg, Deepwalk \cite{perozzi2014deepwalk}, node2vec \cite{grover2016node2vec}, Line\cite{tang2015line}, we are able to learn graph reconstruction by inferring some unobserved transition. To be more specifically, instead of learning $P_\pi(s'|s)$ for a fixed policy $\pi$, we can learn the state-action transition probability $P(s'|s,a)$, which is independent of $\pi$. In this way, we can take use of trajectory data in all history no matter the policy changes or not. And once we are given a policy, we can infer the graph by combining $\pi(a|s)$ and $P(s'|s,a)$.

\subsubsection{Graph Embedding as Auxiliary Task for Representation Learning}
Learning auxiliary tasks is showed to be helpful for state representation learning \cite{jaderberg2016reinforcement}, which is critical to learn a good policy for agents. Among the methods, successor representation is showed to be theoretically and empirically important for learning a good state representations \cite{dayan1993improving,kulkarni2016deep}. Modeling the successor triplet $(s,a,s')$ for MDP is essentially equivalent to modeling the triplet $\left( \text{head}, \text{relation}, \text{tail} \right)$ in knowledge graph. And there exist a lot of algorithms in knowledge graph embedding community to address triplet embedding problem, \eg, TransE \cite{bordes2013translating}, RotateE \cite{sun2019rotate}, QuatE \cite{zhang2019quaternion} and DihEdral \cite{xu2019relation}. These methods can be borrowed to learn richer representation for RL tasks.


\bibliographystyle{abbrv}
\bibliography{references.bib}

\begin{thebibliography}{10}

\bibitem{abu2019mixhop}
S.~Abu-El-Haija, B.~Perozzi, A.~Kapoor, N.~Alipourfard, K.~Lerman,
  H.~Harutyunyan, G.~Ver~Steeg, and A.~Galstyan.
\newblock Mixhop: Higher-order graph convolutional architectures via sparsified
  neighborhood mixing.
\newblock In {\em international conference on machine learning}, pages 21--29.
  PMLR, 2019.

\bibitem{bo2021beyond}
D.~Bo, X.~Wang, C.~Shi, and H.~Shen.
\newblock Beyond low-frequency information in graph convolutional networks.
\newblock {\em arXiv preprint arXiv:2101.00797}, 2021.

\bibitem{bordes2013translating}
A.~Bordes, N.~Usunier, A.~Garcia-Duran, J.~Weston, and O.~Yakhnenko.
\newblock Translating embeddings for modeling multi-relational data.
\newblock In {\em Advances in neural information processing systems}, pages
  2787--2795, 2013.

\bibitem{bronstein2016geometric}
M.~M. Bronstein, J.~Bruna, Y.~LeCun, A.~Szlam, and P.~Vandergheynst.
\newblock Geometric deep learning: Going beyond euclidean data.
\newblock {\em arXiv}, abs/1611.08097, 2016.

\bibitem{chen2020simple}
M.~Chen, Z.~Wei, Z.~Huang, B.~Ding, and Y.~Li.
\newblock Simple and deep graph convolutional networks.
\newblock {\em arXiv preprint arXiv:2007.02133}, 2020.

\bibitem{chien2021adaptive}
E.~Chien, J.~Peng, P.~Li, and O.~Milenkovic.
\newblock Adaptive universal generalized pagerank graph neural network.
\newblock In {\em International Conference on Learning Representations.
  https://openreview. net/forum}, 2021.

\bibitem{chung1997spectral}
F.~R. Chung and F.~C. Graham.
\newblock {\em Spectral graph theory}.
\newblock Number~92. American Mathematical Soc., 1997.

\bibitem{dayan1993improving}
P.~Dayan.
\newblock Improving generalization for temporal difference learning: The
  successor representation.
\newblock {\em Neural Computation}, 5(4):613--624, 1993.

\bibitem{defferrard2016fast}
M.~Defferrard, X.~Bresson, and P.~Vandergheynst.
\newblock Convolutional neural networks on graphs with fast localized spectral
  filtering.
\newblock {\em arXiv}, abs/1606.09375, 2016.

\bibitem{ekambaram2014graph}
V.~N. Ekambaram.
\newblock {\em Graph structured data viewed through a fourier lens}.
\newblock University of California, Berkeley, 2014.

\bibitem{frommer2017block}
A.~Frommer, K.~Lund, and D.~B. Szyld.
\newblock Block {Krylov} subspace methods for functions of matrices.
\newblock {\em Electronic Transactions on Numerical Analysis}, 47:100--126,
  2017.

\bibitem{gilmer2017neural}
J.~Gilmer, S.~S. Schoenholz, P.~F. Riley, O.~Vinyals, and G.~E. Dahl.
\newblock Neural message passing for quantum chemistry.
\newblock In {\em Proceedings of the 34th International Conference on Machine
  Learning-Volume 70}, pages 1263--1272. JMLR. org, 2017.

\bibitem{glorot2010understanding}
X.~Glorot and Y.~Bengio.
\newblock Understanding the difficulty of training deep feedforward neural
  networks.
\newblock In {\em Proceedings of the thirteenth international conference on
  artificial intelligence and statistics}, pages 249--256, 2010.

\bibitem{grover2016node2vec}
A.~Grover and J.~Leskovec.
\newblock node2vec: Scalable feature learning for networks.
\newblock In {\em Proceedings of the 22nd ACM SIGKDD international conference
  on Knowledge discovery and data mining}, pages 855--864. ACM, 2016.

\bibitem{gutknecht2009block}
M.~H. Gutknecht and T.~Schmelzer.
\newblock The block grade of a block krylov space.
\newblock {\em Linear Algebra and its Applications}, 430(1):174--185, 2009.

\bibitem{hamilton2020graph}
W.~L. Hamilton.
\newblock Graph representation learning.
\newblock {\em Synthesis Lectures on Artifical Intelligence and Machine
  Learning}, 14(3):1--159, 2020.

\bibitem{hamilton2017inductive}
W.~L. Hamilton, R.~Ying, and J.~Leskovec.
\newblock Inductive representation learning on large graphs.
\newblock {\em arXiv}, abs/1706.02216, 2017.

\bibitem{he2015delving}
K.~He, X.~Zhang, S.~Ren, and J.~Sun.
\newblock Delving deep into rectifiers: Surpassing human-level performance on
  imagenet classification.
\newblock In {\em Proceedings of the IEEE international conference on computer
  vision}, pages 1026--1034, 2015.

\bibitem{hinton2006fast}
G.~E. Hinton, S.~Osindero, and Y.-W. Teh.
\newblock A fast learning algorithm for deep belief nets.
\newblock {\em Neural computation}, 18(7):1527--1554, 2006.

\bibitem{hou2019measuring}
Y.~Hou, J.~Zhang, J.~Cheng, K.~Ma, R.~T. Ma, H.~Chen, and M.-C. Yang.
\newblock Measuring and improving the use of graph information in graph neural
  networks.
\newblock In {\em International Conference on Learning Representations}, 2019.

\bibitem{hua2022graph}
C.~Hua, S.~Luan, Q.~Zhang, and J.~Fu.
\newblock Graph neural networks intersect probabilistic graphical models: A
  survey.
\newblock {\em arXiv preprint arXiv:2206.06089}, 2022.

\bibitem{jaderberg2016reinforcement}
M.~Jaderberg, V.~Mnih, W.~M. Czarnecki, T.~Schaul, J.~Z. Leibo, D.~Silver, and
  K.~Kavukcuoglu.
\newblock Reinforcement learning with unsupervised auxiliary tasks.
\newblock {\em arXiv preprint arXiv:1611.05397}, 2016.

\bibitem{kingma2014adam}
D.~P. Kingma and J.~Ba.
\newblock Adam: A method for stochastic optimization.
\newblock {\em arXiv preprint arXiv:1412.6980}, 2014.

\bibitem{kipf2016classification}
T.~N. Kipf and M.~Welling.
\newblock Semi-supervised classification with graph convolutional networks.
\newblock {\em arXiv}, abs/1609.02907, 2016.

\bibitem{klicpera2018predict}
J.~Klicpera, A.~Bojchevski, and S.~G{\"u}nnemann.
\newblock Predict then propagate: Graph neural networks meet personalized
  pagerank.
\newblock {\em arXiv preprint arXiv:1810.05997}, 2018.

\bibitem{martin2018diffusion}
M.~Klissarov and D.~Precup.
\newblock Diffusion-based approximate value functions.
\newblock In {\em the 35th international conference on Machine learning ECA
  Workshop}, 2018.

\bibitem{martin2019graph}
M.~Klissarov and D.~Precup.
\newblock Graph convolutional networks as reward shaping functions.
\newblock In {\em ICLR 2019, Representation Learning on Graphs and Manifolds
  Workshop}, 2019.

\bibitem{klissarov2020reward}
M.~Klissarov and D.~Precup.
\newblock Reward propagation using graph convolutional networks.
\newblock {\em arXiv preprint arXiv:2010.02474}, 2020.

\bibitem{kulkarni2016deep}
T.~D. Kulkarni, A.~Saeedi, S.~Gautam, and S.~J. Gershman.
\newblock Deep successor reinforcement learning.
\newblock {\em arXiv preprint arXiv:1606.02396}, 2016.

\bibitem{lecun2015deep}
Y.~LeCun, Y.~Bengio, and G.~Hinton.
\newblock Deep learning.
\newblock {\em nature}, 521(7553):436, 2015.

\bibitem{lecun1998gradient}
Y.~LeCun, L.~Bottou, Y.~Bengio, P.~Haffner, et~al.
\newblock Gradient-based learning applied to document recognition.
\newblock {\em Proceedings of the IEEE}, 86(11):2278--2324, 1998.

\bibitem{li2018deeper}
Q.~Li, Z.~Han, and X.~Wu.
\newblock Deeper insights into graph convolutional networks for semi-supervised
  learning.
\newblock {\em arXiv}, abs/1801.07606, 2018.

\bibitem{li2018adaptive}
R.~Li, S.~Wang, F.~Zhu, and J.~Huang.
\newblock Adaptive graph convolutional neural networks.
\newblock In {\em Thirty-Second AAAI Conference on Artificial Intelligence},
  2018.

\bibitem{liao2019lanczos}
R.~Liao, Z.~Zhao, R.~Urtasun, and R.~S. Zemel.
\newblock Lanczosnet: Multi-scale deep graph convolutional networks.
\newblock {\em arXiv}, abs/1901.01484, 2019.

\bibitem{lim2021new}
D.~Lim, X.~Li, F.~Hohne, and S.-N. Lim.
\newblock New benchmarks for learning on non-homophilous graphs.
\newblock {\em arXiv preprint arXiv:2104.01404}, 2021.

\bibitem{liu2020non}
M.~Liu, Z.~Wang, and S.~Ji.
\newblock Non-local graph neural networks.
\newblock {\em arXiv preprint arXiv:2005.14612}, 2020.

\bibitem{loshchilov2017decoupled}
I.~Loshchilov and F.~Hutter.
\newblock Decoupled weight decay regularization.
\newblock {\em arXiv preprint arXiv:1711.05101}, 2017.

\bibitem{luan2019revisit}
S.~Luan, X.-W. Chang, and D.~Precup.
\newblock Revisit policy optimization in matrix form.
\newblock {\em arXiv preprint arXiv:1909.09186}, 2019.

\bibitem{luan2022we}
S.~Luan, C.~Hua, Q.~Lu, J.~Zhu, X.-W. Chang, and D.~Precup.
\newblock When do we need gnn for node classification?
\newblock {\em arXiv preprint arXiv:2210.16979}, 2022.

\bibitem{luan2021heterophily2}
S.~Luan, C.~Hua, Q.~Lu, J.~Zhu, M.~Zhao, S.~Zhang, X.-W. Chang, and D.~Precup.
\newblock Is heterophily a real nightmare for graph neural networks on
  performing node classification?

\bibitem{luan2021heterophily}
S.~Luan, C.~Hua, Q.~Lu, J.~Zhu, M.~Zhao, S.~Zhang, X.-W. Chang, and D.~Precup.
\newblock Is heterophily a real nightmare for graph neural networks to do node
  classification?
\newblock {\em arXiv preprint arXiv:2109.05641}, 2021.

\bibitem{luan2022revisiting}
S.~Luan, C.~Hua, Q.~Lu, J.~Zhu, M.~Zhao, S.~Zhang, X.-W. Chang, and D.~Precup.
\newblock Revisiting heterophily for graph neural networks.
\newblock {\em Advances in neural information processing systems},
  35:1362--1375, 2022.

\bibitem{luan2023graph}
S.~Luan, C.~Hua, M.~Xu, Q.~Lu, J.~Zhu, X.-W. Chang, J.~Fu, J.~Leskovec, and
  D.~Precup.
\newblock When do graph neural networks help with node classification:
  Investigating the homophily principle on node distinguishability.
\newblock {\em arXiv preprint arXiv:2304.14274}, 2023.

\bibitem{luan2019break}
S.~Luan, M.~Zhao, X.-W. Chang, and D.~Precup.
\newblock Break the ceiling: Stronger multi-scale deep graph convolutional
  networks.
\newblock {\em Advances in neural information processing systems}, 32, 2019.

\bibitem{luan2020training}
S.~Luan, M.~Zhao, X.-W. Chang, and D.~Precup.
\newblock Training matters: Unlocking potentials of deeper graph convolutional
  neural networks.
\newblock {\em arXiv preprint arXiv:2008.08838}, 2020.

\bibitem{luan2020complete}
S.~Luan, M.~Zhao, C.~Hua, X.-W. Chang, and D.~Precup.
\newblock Complete the missing half: Augmenting aggregation filtering with
  diversification for graph convolutional networks.
\newblock {\em arXiv preprint arXiv:2008.08844}, 2020.

\bibitem{luan2022complete}
S.~Luan, M.~Zhao, C.~Hua, X.-W. Chang, and D.~Precup.
\newblock Complete the missing half: Augmenting aggregation filtering with
  diversification for graph convolutional neural networks.
\newblock {\em arXiv preprint arXiv:2212.10822}, 2022.

\bibitem{maehara2019revisiting}
T.~Maehara.
\newblock Revisiting graph neural networks: All we have is low-pass filters.
\newblock {\em arXiv preprint arXiv:1905.09550}, 2019.

\bibitem{mahadevan2005proto}
S.~Mahadevan.
\newblock Proto-value functions: Developmental reinforcement learning.
\newblock In {\em Proceedings of the 22nd international conference on Machine
  learning}, pages 553--560. ACM, 2005.

\bibitem{monti2017geometric}
F.~Monti, D.~Boscaini, J.~Masci, E.~Rodola, J.~Svoboda, and M.~M. Bronstein.
\newblock Geometric deep learning on graphs and manifolds using mixture model
  cnns.
\newblock In {\em Proceedings of the IEEE Conference on Computer Vision and
  Pattern Recognition}, pages 5115--5124, 2017.

\bibitem{pei2020geom}
H.~Pei, B.~Wei, K.~C.-C. Chang, Y.~Lei, and B.~Yang.
\newblock Geom-gcn: Geometric graph convolutional networks.
\newblock {\em arXiv preprint arXiv:2002.05287}, 2020.

\bibitem{perozzi2014deepwalk}
B.~Perozzi, R.~Al-Rfou, and S.~Skiena.
\newblock Deepwalk: Online learning of social representations.
\newblock In {\em Proceedings of the 20th ACM SIGKDD international conference
  on Knowledge discovery and data mining}, pages 701--710. ACM, 2014.

\bibitem{raghavan2007near}
U.~N. Raghavan, R.~Albert, and S.~Kumara.
\newblock Near linear time algorithm to detect community structures in
  large-scale networks.
\newblock {\em Physical review E}, 76(3):036106, 2007.

\bibitem{ribeiro2017struc2vec}
L.~F. Ribeiro, P.~H. Saverese, and D.~R. Figueiredo.
\newblock struc2vec: Learning node representations from structural identity.
\newblock In {\em Proceedings of the 23rd ACM SIGKDD international conference
  on knowledge discovery and data mining}, pages 385--394, 2017.

\bibitem{sun2019rotate}
Z.~Sun, Z.-H. Deng, J.-Y. Nie, and J.~Tang.
\newblock Rotate: Knowledge graph embedding by relational rotation in complex
  space.
\newblock {\em arXiv preprint arXiv:1902.10197}, 2019.

\bibitem{sutton2018reinforcement}
R.~S. Sutton and A.~G. Barto.
\newblock {\em Reinforcement learning: An introduction}.
\newblock MIT press, 2018.

\bibitem{tang2015line}
J.~Tang, M.~Qu, M.~Wang, M.~Zhang, J.~Yan, and Q.~Mei.
\newblock Line: Large-scale information network embedding.
\newblock In {\em Proceedings of the 24th international conference on world
  wide web}, pages 1067--1077. International World Wide Web Conferences
  Steering Committee, 2015.

\bibitem{velivckovic2017attention}
P.~Velickovic, G.~Cucurull, A.~Casanova, A.~Romero, P.~Lio, and Y.~Bengio.
\newblock Graph attention networks.
\newblock {\em arXiv}, abs/1710.10903, 2017.

\bibitem{wang2007dual}
T.~Wang, M.~Bowling, and D.~Schuurmans.
\newblock Dual representations for dynamic programming and reinforcement
  learning.
\newblock In {\em 2007 IEEE International Symposium on Approximate Dynamic
  Programming and Reinforcement Learning}, pages 44--51. IEEE, 2007.

\bibitem{wu2019simplifying}
F.~Wu, T.~Zhang, A.~H.~d. Souza~Jr, C.~Fifty, T.~Yu, and K.~Q. Weinberger.
\newblock Simplifying graph convolutional networks.
\newblock {\em arXiv preprint arXiv:1902.07153}, 2019.

\bibitem{wu2019survey}
Z.~Wu, S.~Pan, F.~Chen, G.~Long, C.~Zhang, and P.~S. Yu.
\newblock A comprehensive survey on graph neural networks.
\newblock {\em arXiv}, abs/1901.00596, 2019.

\bibitem{xu2019relation}
C.~Xu and R.~Li.
\newblock Relation embedding with dihedral group in knowledge graph.
\newblock {\em arXiv preprint arXiv:1906.00687}, 2019.

\bibitem{pmlr-v80-xu18c}
K.~Xu, C.~Li, Y.~Tian, T.~Sonobe, K.-i. Kawarabayashi, and S.~Jegelka.
\newblock Representation learning on graphs with jumping knowledge networks.
\newblock In J.~Dy and A.~Krause, editors, {\em Proceedings of the 35th
  International Conference on Machine Learning}, volume~80 of {\em Proceedings
  of Machine Learning Research}, pages 5453--5462. PMLR, 10--15 Jul 2018.

\bibitem{yan2021two}
Y.~Yan, M.~Hashemi, K.~Swersky, Y.~Yang, and D.~Koutra.
\newblock Two sides of the same coin: Heterophily and oversmoothing in graph
  convolutional neural networks.
\newblock {\em arXiv preprint arXiv:2102.06462}, 2021.

\bibitem{zhang2019quaternion}
S.~Zhang, Y.~Tay, L.~Yao, and Q.~Liu.
\newblock Quaternion knowledge graph embedding.
\newblock {\em arXiv preprint arXiv:1904.10281}, 2019.

\bibitem{zhang2018graph}
S.~Zhang, H.~Tong, J.~Xu, and R.~Maciejewski.
\newblock Graph convolutional networks: Algorithms, applications and open
  challenges.
\newblock In {\em International Conference on Computational Social Networks},
  pages 79--91. Springer, 2018.

\bibitem{zhao2021consciousness}
M.~Zhao, Z.~Liu, S.~Luan, S.~Zhang, D.~Precup, and Y.~Bengio.
\newblock A consciousness-inspired planning agent for model-based reinforcement
  learning.
\newblock {\em Advances in neural information processing systems},
  34:1569--1581, 2021.

\bibitem{zhao2019meta}
M.~Zhao, S.~Luan, I.~Porada, X.-W. Chang, and D.~Precup.
\newblock Meta-learning state-based eligibility traces for more
  sample-efficient policy evaluation.
\newblock {\em arXiv preprint arXiv:1904.11439}, 2019.

\bibitem{zhu2020graph}
J.~Zhu, R.~A. Rossi, A.~Rao, T.~Mai, N.~Lipka, N.~K. Ahmed, and D.~Koutra.
\newblock Graph neural networks with heterophily.
\newblock {\em arXiv preprint arXiv:2009.13566}, 2020.

\bibitem{zhu2020beyond}
J.~Zhu, Y.~Yan, L.~Zhao, M.~Heimann, L.~Akoglu, and D.~Koutra.
\newblock Beyond homophily in graph neural networks: Current limitations and
  effective designs.
\newblock {\em Advances in Neural Information Processing Systems}, 33, 2020.

\bibitem{zhu2020generalizing}
J.~Zhu, Y.~Yan, L.~Zhao, M.~Heimann, L.~Akoglu, and D.~Koutra.
\newblock Generalizing graph neural networks beyond homophily.
\newblock {\em arXiv preprint arXiv:2006.11468}, 2020.

\end{thebibliography}
\newpage

\appendix
\section{Calculation of Variances}
\label{appendix}
\subsection{Background}

We first decompose the deep GCN architecture as follows
\begin{equation}\label{eq:deep_gcn_decompose_full}
\begin{aligned}
   & \bm{Y_0} = \bm{X}, \;\bm{H_1} = \hat{A} \bm{X} W_0, \; \bm{Y_1} = f (\bm{H_1}) \\ 
   & \bm{H_{l+1}}  = \hat{A} \bm{Y_l} W_l, \; \bm{Y_{l+1}} = f (\bm{H_{l+1}}), \; l = 1, \dots, n\\
   & \bm{Y} = \text{softmax} (\hat{A}  \bm{Y_n} W_n ) \equiv  \text{softmax} (\bm{H_{n+1}}) \\
   & L = -\text{trace} (\bm{Z}^T \text{log} \bm{Y})\\
\end{aligned}
\end{equation}

where $\bm{H_l},\bm{Y_l} \in \mathbb{R}^{N \times F_{l}}$, $W_l \in \mathbb{R}^{F_l \times F_{l+1}}$;  $\bm{Z}\in\Rbb^{N\times C}$ is the ground truth matrix with one-hot label vector $\bm{Z}_{i,:}$ in each row, $C$ is number of classes; $L$ is the scalar loss. Then the gradient propagates in the following way 
\begin{equation}\label{eq:gradient_appendix}
\begin{aligned}
   \text{Output} \;& \frac{\partial L }{\partial \bm{H_{n+1}}} = \text{softmax}(\bm{H_{n+1}}) - \bm{Z} \\ 
   &\frac{\partial L}{\partial W_n} = \bm{Y_n}^T \hat{A} \frac{\partial L}{ \partial \bm{H_{n+1}}}, \; \frac{\partial L}{\partial \bm{Y_n}} = \hat{A} \frac{\partial L}{\partial \bm{H_{n+1}}}  W_n^T\\
   \text{Hidden} \;& \frac{\partial L}{\partial \bm{H_l}} = \frac{\partial L }{\partial \bm{Y_l}} \odot f'(\bm{H_l}), \;
   \frac{\partial L}{\partial W_{l-1}} = \bm{Y_{l-1}}^T \hat{A} \frac{\partial L }{\partial \bm{H_l} }, \\
   &\frac{\partial L}{\partial \bm{Y_{l-1}}} = \hat{A} \frac{\partial L}{ \partial \bm{H_l}}  W_{l-1}^T  \\
\end{aligned}
\end{equation}
where $\odot$ is the Hadamard product. The gradient propagation of GCN differs from that of multi-layer perceptron (MLP) by an extra multiplication of $\hat{A}$ when the gradient signal flows through $\bm{Y_l}$. 

\subsection{Variance Analysis}

\subsubsection{Forward View}
\label{appendix:forward_view}
\begin{equation}
\begin{aligned}
    & \bm{H_{l+1}} = \hat{A} \bm{Y_l} W_l, \; \left(\bm{H_{l+1}}\right)_{ij} = \hat{A}_{i,:} \bm{Y_l} (W_l)_{:,j} = \sum\limits_{t=1}^{F_l} \sum\limits_{k=1}^N \hat{A}_{ik} \left(\bm{Y_l}\right)_{kt} (W_l)_{t,j} \\
    &\bm{Y_{l+1}} = f (\bm{H_{l+1}}), \; l = 1, \dots, n 
\end{aligned}
\end{equation}
Suppose the activation function is identity function such as \cite{wu2019simplifying}, all element in $W$ share the same variance and each element in $\bm{X}$ are independent and share the same variance, $E\left((\bm{Y_l})_{kt}\right) = 0$ and $E\left((W_l)_{ij}\right) = 0$, $\text{Var}\left((\bm{Y_{l+1}})_{ij}\right)$ can be written as
\begin{equation}
\begin{aligned} \label{eq:variance_hidden_units2}
   & \text{Var}\left(\sum\limits_{t=1}^{F_l} \sum\limits_{k=1}^N \hat{A}_{ik} \left(\bm{Y_l}\right)_{kt} W_{t,j}\right) = \sum\limits_{t=1}^{F_l} \sum\limits_{k=1}^N \text{Var}\left( \hat{A}_{ik} \left(\bm{Y_l}\right)_{kt} W_{t,j}\right)\\
   & = F_l (d_i+1) \cdot \frac{1}{(d_i+1)^2} \text{Var}\left(\bm{Y_l}\right) \text{Var}\left(W\right) \\
   & = \frac{F_l}{d_i+1} \text{Var}\left(\bm{Y_l}) \text{Var}(W\right) = \text{Var}\left(\bm{Y_{l+1}}\right)\\
\end{aligned}
\end{equation}
Then, if we want $\text{Var}\left(\bm{Y_{l+1}}\right) = \text{Var} \left(\bm{Y_{l}}\right)$ we will have
\begin{equation}
    \text{Var}(W) = \frac{d_i+1}{F_l}
\end{equation}

Since the parameter matrix is shared by all nodes, we cannot design a node specified initialization scheme for the $W$. Thus, we initialize each element in $W$ by the average values as follows
\begin{equation}
    \text{Var}(W) = \frac{\sum\limits_{i=1}^N (d_i+1)}{N F_l} = \frac{1+\text{average node degree}}{F_l}
\end{equation}

If we relax the assumption $E\left((\bm{Y_l})_{kt}\right) = 0$ and assume it nonzero, as shown in \cite{he2015delving}, if we use ReLU activation function, we will have 

\begin{equation}
    \text{Var}(W) = \frac{2\sum\limits_{i=1}^N (d_i+1)}{N F_l} = \frac{2(1+\text{average node degree})}{F_l}
\end{equation}

If we consider the correlation between nodes and keep the assumption that each dimension of the feature (columns in $\bm{X}$) is independent and $E(X) \neq 0$, we will have
\begin{equation}
\begin{aligned}
   & \text{Var}\left((\bm{H_{l+1}})_{ij}\right) = \text{Var}\left(\sum\limits_{t=1}^{F_l} \sum\limits_{k=1}^N \hat{A}_{ik} \left(\bm{Y_l}\right)_{kt} W_{t,j}\right) \\
   & = F_l \cdot \text{Var}\left(\left(\sum\limits_{k=1}^N A_{ik} \left(\bm{Y_l}\right)_{k} \right) W \right) = F_l \cdot E\left( \sum\limits_{k=1}^N  A_{ik} \left(\bm{Y_l}\right)_{k} \right)^2 \text{Var}(W) \\
   & = \frac{F_l }{(d_i+1)^2} E\left( \sum\limits_{k\in\mathcal{N}_i}    \left(\bm{Y_l}\right)_{k} \right)^2 \text{Var}(W)  \\
   & = \frac{F_l}{(d_i+1)^2} E\left( \sum\limits_{k\in\mathcal{N}_i} \left(\bm{Y_l}\right)_{k}^2 + \sum\limits_{k,j \in\mathcal{N}_i, k \neq j} \left(\bm{Y_l}\right)_{k} \left(\bm{Y_l}\right)_{j} \right) \text{Var}(W) \\
\end{aligned}
\end{equation}

Since 
$$E\left(\left(\bm{Y_l}\right)_{k} \left(\bm{Y_l}\right)_{j}\right) = \text{Cov}\left(\left(\bm{Y_l}\right)_{k}, \left(\bm{Y_l}\right)_{j}\right) + E\left(\left(\bm{Y_l}\right)_{k}\right) E\left(\left(\bm{Y_l}\right)_{j}\right)$$
we can have several reasonable assumptions over $\text{Cov}\left(\left(\bm{Y_l} \right)_{k}, \left(\bm{Y_l}\right)_{j}\right)$ to get different results
\begin{itemize}
    \item  The adjacency matrix with self-loop can be considered as a prior covariance matrix and thus a reasonable assumption is $\text{Cov}\left(\left(\bm{Y_l} \right)_{k}, \left(\bm{Y_l}\right)_{j}\right) = \text{Var}\left(\left(\bm{Y_l} \right)_{k}\right) = \text{Var}\left(\left(\bm{Y_l} \right)_{j}\right)$.
     \item Consider symmetric normalized $\hat{A}$ as a prior covariance matrix and we have $\text{Cov}\left(\left(\bm{Y_l} \right)_{k}, \left(\bm{Y_l}\right)_{j}\right) = \sqrt{\text{Var}\left(\left(\bm{Y_l} \right)_{k}\right) \text{Var}\left(\left(\bm{Y_l} \right)_{j}\right)} = \text{Var}\left(\left(\bm{Y_l} \right)_{k}\right)$
\end{itemize}
These assumptions all lead us to
$$E\left(\left(\bm{Y_l}\right)_{k} \left(\bm{Y_l}\right)_{j}\right) = \text{Var}\left(\left(\bm{Y_l} \right)_{k}\right) + E^2\left(\left(\bm{Y_l} \right)_{k}\right) = E\left(\bm{Y_l} \right)_{k}^2$$
   
Thus we have 
\begin{equation}
\begin{aligned}
   & \text{Var}\left((\bm{H_{l+1}})_{ij}\right) = F_l \cdot \frac{1}{(d_i+1)^2} E\left( (d_i+1)^2 \left(\bm{Y_l} \right)_{k}^2 \right) \text{Var}(W) \\
   & = F_l \cdot E\left(\left(\bm{Y_l} \right)_{k}^2 \right) \text{Var}(W) = F_l \cdot \frac{1}{2} \text{Var}\left(\bm{H}_{l} \right) \text{Var}(W) \\
\end{aligned}
\end{equation}
Thus

\begin{equation}
\begin{aligned}
  \text{Var}(W) = \frac{2}{F_l} \\
\end{aligned}
\end{equation}

Suppose $E(\left(\bm{Y_l} \right)_{k} \left(\bm{Y_l} \right)_{j}) \geq 0$ and $Cov(\left(\bm{Y_l} \right)_{k},\left(\bm{Y_l} \right)_{j}) \geq 0$, since $Cov(\left(\bm{Y_l} \right)_{k}, \left(\bm{Y_l} \right)_{j}) \leq \sqrt{\text{Var}(\left(\bm{Y_l} \right)_{k}) \text{Var}(\left(\bm{Y_l} \right)_{j})} = \text{Var}(\left(\bm{Y_l} \right)_{k})$, we have $E(\left(\bm{Y_l} \right)_{k} \left(\bm{Y_l} \right)_{j}) \leq E\left(\bm{Y_l} \right)_{k}^2$ and we assume $E(\left(\bm{Y_l} \right)_{k} \left(\bm{Y_l} \right)_{j}) = \alpha E\left(\bm{Y_l} \right)_{k}^2, \alpha \in [0,1]$. Then,

\begin{equation}
\begin{aligned}
   & \text{Var}\left((\bm{H_{l+1}})_{ij}\right) = F_l \cdot \frac{1}{(d_i+1)^2} (d_i+1)(\alpha d_i+1) E\left(\bm{Y_l} \right)_{k}^2 \text{Var}(W) \\
   & = F_l \cdot \frac{\alpha d_i+1}{d_i+1} E\left(\bm{Y_l} \right)_{k}^2 \text{Var}(W)\\
   & = F_l \cdot \frac{\alpha d_i+1}{2(d_i+1)} \text{Var}\left(\bm{H_{l}} \right) \text{Var}(W)\\
\end{aligned}
\end{equation}
Thus,
\begin{equation}
\begin{aligned}
   & \text{Var}(W) = \frac{2(d_i+1)}{F_l(\alpha d_i+1)}\\
\end{aligned}
\end{equation}

$(d_i+1)/(\alpha d_i+1)$ can be considered as the effective degree of node $i$ and an estimation is
\begin{equation}
\begin{aligned}
    \hat{\text{Var}}(W) & = \frac{\sum_i 2(d_i+1)/(\alpha d_i+1)}{N F_l} \\
   & = \frac{2\times \text{average effective node degree}}{F_l}\\
\end{aligned}
\end{equation}

\subsubsection{Backward View}
\label{appendix:backward_view}
Under linear assumption, we have $\bm{H_l} = \bm{Y_l}$ and

\begin{equation}\label{eq:backward_view2}
\begin{aligned}
  \frac{\partial L}{\partial \bm{H_l}} = \frac{\partial L }{\partial \bm{Y_l}} = \hat{A} \frac{\partial L}{ \partial \bm{H_{l+1}}}  W_{l}^T  , \;
   \frac{\partial L}{\partial W_{l-1}} = \bm{Y_{l-1}}^T \hat{A} \frac{\partial L }{\partial \bm{H_l} }, \\
\end{aligned}
\end{equation}
We have
\begin{equation}
\begin{aligned}
   \left(\frac{\partial L}{\partial \bm{H_l}}\right)_{ij} & = \sum\limits_{t=1}^{F_{l+1}} \sum\limits_{k=1}^N \hat{A}_{ik} \left(\frac{\partial L}{\partial \bm{H_{l+1}}}\right)_{kt} (W_l^T)_{t,j} \\
   \left(\frac{\partial L}{\partial W_{l-1}}\right)_{ij} & = (\hat{A} \bm{Y_{l-1}})_{\cdot i}^T  \left(\frac{\partial L}{\partial \bm{H_l} }\right)_{\cdot j} = \sum\limits_{k=1}^N  (\sum\limits_{t=1}^N \hat{A}_{kt} (\bm{Y_{l-1}})_{t i}) \left(\frac{\partial L}{\partial \bm{H_l} }\right)_{k j}, \\
   \end{aligned}
\end{equation}
And 
\begin{equation}
\begin{aligned}
   \text{Var}\left(\left(\frac{\partial L}{\partial \bm{H_l}}\right)_{ij}\right) & = \text{Var}\left( \sum\limits_{t=1}^{F_{l+1}} \sum\limits_{k=1}^N \hat{A}_{ik} \left(\frac{\partial L}{\partial \bm{H_{l+1}}}\right)_{kt} (W_l^T)_{t,j} \right)\\
   & = \frac{F_{l+1}}{d_i+1}\text{Var}\left(\left(\frac{\partial L}{\partial \bm{H_{l+1}}}\right)_{kt} (W_l^T)_{t,j} \right)\\
   & = \frac{F_{l+1}}{d_i+1} \text{Var}\left( \frac{\partial L}{\partial \bm{H_{l+1}}}  \right) \text{Var}\left(W_l\right)\\
   & \approx \frac{N F_{l+1}}{\sum_i(d_i+1)} \text{Var}\left( \frac{\partial L}{\partial \bm{H_{l+1}}}  \right) \text{Var}\left(W_l\right)\\
   & \approx \text{Var}\left(\frac{\partial L}{\partial \bm{H_{n+1}}}  \right) \prod\limits_{l'=l}^n \frac{N F_{l+1}}{\sum_i(d_i+1)} \text{Var}\left(W_l\right)\\
    \text{Var}\left(\left(\frac{\partial L}{\partial W_{l-1}}\right)_{ij}\right) & = \text{Var}\left( \sum\limits_{k=1}^N  (\sum\limits_{t=1}^N \hat{A}_{kt} (\bm{Y_{l-1}})_{t i}) \left(\frac{\partial L}{\partial \bm{H_l} }\right)_{k j} \right) \\
    & = \sum\limits_{k=1}^N \text{Var}\left(\sum\limits_{t=1}^N \hat{A}_{kt} (\bm{Y_{l-1}})_{t i} \right) \text{Var} \left( \left(\frac{\partial L}{\partial \bm{H_l} }\right)_{k j} \right) \\
    & = \left(\sum\limits_{k=1}^N \frac{1}{d_k+1} \right)\text{Var}\left(\bm{Y_{l-1}} \right) \text{Var} \left(\frac{\partial L}{\partial \bm{H_l} } \right) \\
   \end{aligned}
\end{equation}

From \eqref{eq:variance_hidden_units2}, $\text{Var}\left(\bm{Y_{l-1}} \right)$ can be approximately written as
\begin{equation}
\begin{aligned}  
   & \text{Var}\left(\bm{Y_{l-1}}\right)  = \text{Var}\left(\bm{H_{l-1}}\right) \approx \frac{N F_{l-2}}{\sum_k d_k+1} \text{Var}\left(\bm{Y_{l-2}}) \text{Var}(W\right)\\
   & \approx \text{Var}(\bm{Y_{0}}) \prod\limits_{l'=0}^{l-2}\frac{N F_{l'}}{\sum_k d_k+1}  \text{Var}(W)\\
\end{aligned}
\end{equation}

Then

\begin{equation}
\begin{aligned}
    & \text{Var}\left(\left(\frac{\partial L}{\partial W_{l-1}}\right)_{ij}\right) \\ &\approx \left(\sum\limits_{k=1}^N \frac{1}{d_k+1} \right)\text{Var}(\bm{Y_{0}}) \text{Var}\left(\frac{\partial L}{\partial \bm{H_{n+1}}}  \right) \prod\limits_{l'=0, l' \neq l-1}^{n}\frac{N F_{l'}}{\sum_k d_k+1} \text{Var}(W)\\
   \end{aligned}
\end{equation}

\subsection{Energy Analysis}
Another way to design weight initialization is from the flow of energy perspective. Under the linear assumption and suppose we can do QR factorization of the weight matrices to make them orthogonal and $W^TW=WW^T = \alpha I$, then we have
\begin{equation}\label{eq:energy_1}
\begin{aligned}
  & \text{trace}(\bm{H_{l+1}}^T \bm{H_{l+1}}) = \text{trace}\left((\hat{A} \bm{Y_l} W_l)^T \hat{A} \bm{Y_l} W_l \right)\\
  & = \text{trace}\left(W_l^T \bm{Y_l}^T \hat{A}^T \hat{A} \bm{Y_l} W_l \right) = \text{trace}\left(W_l W_l^T \bm{Y_l}^T \hat{A}^T \hat{A} \bm{Y_l} \right) \\
  & = \text{trace}\left(\alpha \sum\limits_{i=1}^N \lambda_i^2 \bm{Y_l}^T \bm{u_i} \bm{u_i^T} \bm{Y_l} \right) = \text{trace}\left(\alpha \sum\limits_{i=1}^N \lambda_i^2 \bm{u_i^T} \bm{Y_l} (\bm{u_i^T} \bm{Y_l})^T \right) \\
  & = \text{trace}\left(\alpha \sum\limits_{i=1}^N \lambda_i^2 \norm{\bm{u_i^T} \bm{Y_l}}_2^2 \right) = \text{trace}\left(\sum\limits_{i=1}^N \norm{\bm{u_i^T} \bm{H_{l+1}}}_2^2 \right) \\
\end{aligned}
\end{equation}

Suppose all $\norm{\bm{u_i^T} \bm{Y_l}}_2^2$ and $\bm{u_i^T} \bm{H_{l+1}}$ are equal, then 
\begin{equation}\label{eq:energy_2}
\begin{aligned}
  \alpha = \frac{N}{\sum\limits_{i=1}^N \lambda_i^2} = \frac{N}{\norm{\hat{A}}_F^2} = \frac{N}{\sum\limits_{i=1}^N\frac{1}{d_i+1}}
\end{aligned}
\end{equation}

If we use ReLU activation function, we have
\begin{equation}\label{eq:energy_3}
\begin{aligned}
  \alpha = \frac{2N}{\sum\limits_{i=1}^N \lambda_i^2} = \frac{2N}{\norm{\hat{A}}_F^2} = \frac{2N}{\sum\limits_{i=1}^N\frac{1}{d_i+1}}
\end{aligned}
\end{equation}

\end{document}